# Identity-Duplication Auditing in National-Scale Neuroimaging Repositories

Jiheng Li*[a], Michael E. Kim[a], Trent M. Schwartz[b], Yuhan Cui[c], Gaurav Rudravaram[b], Derek B. Archer[d], Timothy J. Hohman[d], Lori L. Beason-Held[e], Victoria L. Morgan[f], Dario J. Englot[a,b,f], Angela L. Jefferson[d], for the Alzheimer's Disease Neuroimaging Initiative[1], for the BIOCARD Study team[2], for the Health and Aging Brain Study: Health Disparities (HABS-HD) Study Team[3], Lianrui Zuo[b], Guray Erus[c], Christos Davatzikos[c], Bennett A. Landman[a,b,f]

[a]Department of Computer Science, Vanderbilt University, Nashville, TN, USA; [b]Department of Electrical and Computer Engineering, Vanderbilt University, Nashville, TN, USA; [c]AI2D Center for AI and Data Science for Integrated Diagnostics, University of Pennsylvania, Philadelphia, PA, USA; [d]Vanderbilt Memory and Alzheimer's Center, Vanderbilt Health, Nashville, TN, USA; [e]Laboratory of Behavioral Neuroscience, National Institute on Aging, National Institutes of Health, Baltimore, MD, USA; [f]Department of Radiology and Radiological Sciences, Vanderbilt Health, Nashville, TN, USA; *Corresponding author: jiheng.li.1@vanderbilt.edu

## ABSTRACT

**Objective.** National-scale magnetic resonance imaging (MRI) repositories increasingly integrate data from different studies and institutions. However, subject identifiers that are valid only within individual datasets are no longer guaranteed to remain globally unique after aggregation, making it possible for the same subject to be assigned multiple identifiers, which we define as identity duplication. Such duplication can create leakage between training and test data and inflate apparent performance in downstream biomedical studies. Existing methods do not provide an end-to-end, image-based workflow for auditing this problem at repository scale.

**Methods.** HAPPEN is a human-in-the-loop pipeline for auditing identity duplication in T1-weighted brain MRI repositories. It combines SHA-256 fingerprinting for exact-duplicate detection with supervised contrastive retrieval of non-identical scans that may originate from the same person. Retrieved pairs are reviewed as candidates in a locally hosted interface rather than automatically classified as duplicates. We benchmarked retrieval against PCA and 3D-SIFT-Rank in an internal test set and two withheld cohorts. We deployed the workflow in a 95,129-scan aggregated repository and assessed end-to-end recovery using 54 genetic-reference pairs. Transferability was assessed by locally deploying the same workflow on 22,386 scans at an independent institution without model retraining or image transfer.

---

[1] Data used in preparation of this article were obtained from the Alzheimer's Disease Neuroimaging Initiative (ADNI) database (adni.loni.usc.edu). As such, the investigators within the ADNI contributed to the design and implementation of ADNI and/or provided data but did not participate in analysis or writing of this report. A complete listing of ADNI investigators can be found at: http://adni.loni.usc.edu/wp-content/uploads/how_to_apply/ADNI_Acknowledgement_List.pdf

[2] Data used in preparation of this article were derived from BIOCARD study data, supported by grant U19-AG033655 from the National Institute on Aging. The BIOCARD study team did not participate in the analysis or writing of this report; however, they contributed to the design and implementation of the study. A listing of BIOCARD investigators may be accessed at: https://biocard.pathology.jhu.edu/our-team/

[3] HABS-HD MPIs and Investigators are listed in the Acknowledgements.

**Results.** HAPPEN achieved near-ceiling same-subject retrieval across all three cohorts and matched or exceeded the baseline point estimates on every metric. Deployment in the study repository identified 1,316 exact-duplicate scan groups and 1,275 reviewer-supported near-duplicate subject groups. Of these groups, 56% and 82%, respectively, crossed dataset boundaries. All 54 genetic-reference pairs were recovered. The external team independently completed the full workflow using a locally selected operating threshold and review standard.

**Conclusion.** Together, these findings indicate that identity duplication is a structural risk of MRI repository aggregation and that HAPPEN can support practical, locally governed quality assurance.

***Keywords--*** identity duplication; neuroimaging repositories; data leakage; quality assurance; contrastive learning

## 1. INTRODUCTION

To improve statistical power and reduce estimation uncertainty associated with limited sample sizes, an increasing number of neuroimaging studies are moving toward large-scale dataset aggregation [1], [2], [3], [4]. Magnetic resonance imaging (MRI) scans from different studies, sites, years, and scanners have been pooled into national-level repositories. Studies typically assign de-identified subject identifiers to participants for privacy protection during data collection, and subject identifiers in a single dataset are often only valid within the scope of that dataset [5]. However, when data from multiple studies and sites are merged, there is no guarantee that these "dataset-specific" identifiers remain globally valid, and it is even possible for the same individual to be assigned multiple identifiers. In this paper, this "one-subject, multiple-identifiers" phenomenon is termed *identity duplication*.

Identity duplication can arise from a highly complex set of pathways, many of which are side effects of routine data-governance processes (Figure 1). First, within a single study, researchers may re-issue new identifiers for existing participants. This commonly occurs when a study is updated and legacy records coexist with newly released records, or when the study itself compiles participants drawn from several ongoing projects [6]. Issuing new identifiers is itself a legitimate data-governance practice because it can help fully decouple the identifiers used in a public release from potentially identifiable information [7], [8]. However, when repository maintainers or downstream researchers combine cohorts or release versions, the same individual may appear under both legacy and new identifiers, creating duplication in the aggregated dataset. Second, identity duplication can also arise across studies and across sites [9], [10], [11], [12]. One possible explanation is dual enrollment. Shiovitz et al. describe "professional" participants who may enroll in multiple studies and travel across sites [13]; they report a case in which one individual was screened at a minimum of seven different sites in 12 months [14]. Another pathway is data transfer across studies and across sites. Data sharing and transfer are widespread in neuroimaging [15], but large-scale transfers themselves are inherently risky: beyond potential miscommunication during handoffs, duplication created for the reasons above can be inadvertently copied and propagated through repeated transfers and re-aggregation, spreading into studies and sites that originally had no duplication. Overall, most of these causes are well-intentioned design choices made for privacy protection, version control, and release management. Identity duplication is therefore best understood as a structural risk that emerges when otherwise valid data-governance processes interact at repository scale.

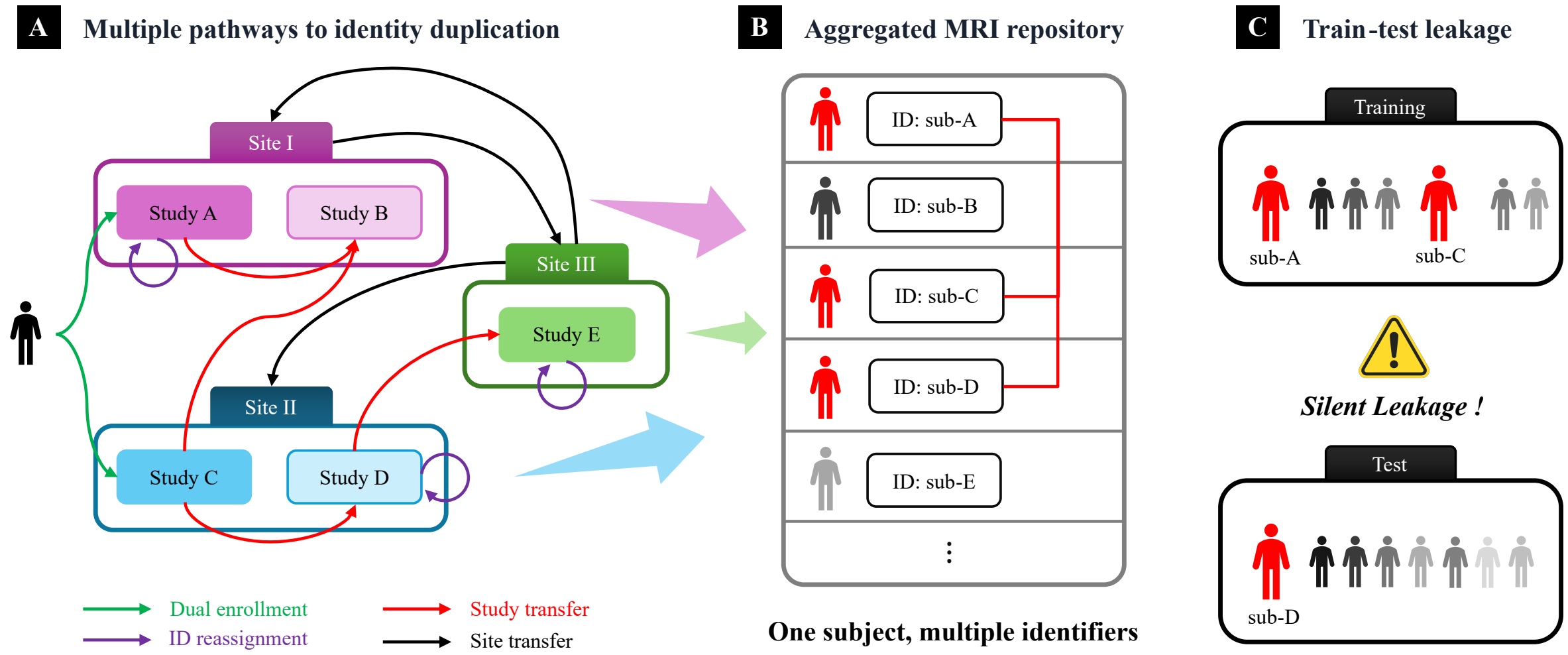

**Figure 1. Identity duplication as a structural risk in aggregated neuroimaging repositories.** Scans from the same subject can enter an aggregated repository through multiple pathways (e.g., participation in multiple studies, reassignment with new identifiers across releases within a single study, and data transfers across studies or sites), resulting in the same person being recorded under different identifiers. If the aggregated data are subsequently analyzed according to the recorded identifiers, identity duplication can increase false-positive risk and create silent train-test leakage. This leakage can inflate apparent performance and compromise downstream conclusions.

Identity duplication can seriously compromise the validity of research based on large-scale neuroimaging repositories [16], [17], [18]. First, when repeated records from the same individual are treated as distinct participants, correlated observations may be analyzed as if they were independent, increasing the risk of false-positive findings [16]. Second, identity duplication can create subject-level leakage in predictive modeling. If duplicated identifiers place scans from one individual in both the training and test sets, the model is partly evaluated on a person it has already seen, so its performance no longer reflects generalization to new participants [17], [19]. Yagis et al. showed that this inflation can be substantial: on a randomly labeled OASIS-derived dataset, slice-level splitting produced test accuracy above 93%, whereas subject-level splitting yielded approximately 50% [19]. Rosenblatt et al. similarly found that repeated subjects inflated connectome-based predictions across age, matrix reasoning, and attention phenotypes [17]. Beyond inflating benchmark performance, such leakage can bias estimates of biomarker performance and make reported biomarkers less reliable in new data [18]. A recent review of Alzheimer's disease classification further linked recurrent leakage to the gap between reported accuracy and clinical translation [20]. Thus, identity duplication can make a model appear to have learned meaningful biological information when it is partly relying on repeated individuals, weakening confidence in downstream biomedical conclusions [18], [20], [21].

As identity duplication poses a serious risk to neuroimaging data science, practical methods are needed to identify it in aggregated datasets. This study introduces HAPPEN (Human-in-the-loop Auditing Pipeline for Exact and Near Duplicates), an end-to-end workflow for auditing identity duplication. First, it uses SHA-256 fingerprints to identify scans with the same image content, termed *exact duplicates*. Second, it uses a learned MRI identity model to retrieve scans that are not identical but may come from the same person, termed *near-duplicate candidates*. These candidates are then assessed through human review. HAPPEN was developed using T1-weighted brain MRI from a national-scale aggregated repository. The complete HAPPEN workflow was evaluated through deployment across all 95,129 scans in that repository and subsequently deployed at the University of Pennsylvania to audit an external repository without model retraining.

## 2. RELATED WORK

Although identity duplication has a severe impact, existing work only partially addresses this problem. At the data-governance layer, several national programs provide identifier-linkage systems across studies and sites. For example, the NIMH Data Archive developed a repository-wide Global Unique Identifier (GUID) system to resolve mismatched namespaces across sub-projects and make subject identifiers globally unique within the archive [9], [10]. Similarly, the National Alzheimer's Coordinating Center (NACC) uses the NACCID to link all available data types for an individual participant, including data from partner repositories and affiliated studies [11], [12]. In UK Biobank, when researchers apply for access, each participant is assigned a project-specific encrypted identifier (EID) that is randomly generated and unique within that access application [7], [8]. However, these mechanisms are typically scoped to a particular project space; when we aggregate data across projects and repositories, there is still no unified subject-identifier namespace, and identity duplication cannot be avoided.

At the methodological layer, prior work has established that T1-weighted MRI contains stable subject-specific information that can be used to match repeated scans. BrainPrint used structural brain morphology for subject identification [22], DeepBrainPrint learned representations for rapid same-subject retrieval [23], and Chauvin et al. used 3D keypoint signatures to identify same-subject pairs and previously unknown labeling inconsistencies across four public repositories [24]. These studies establish that MRI can support subject matching and data curation. HAPPEN targets repository-wide identity auditing under a key constraint: recorded subject identifiers cannot be assumed to be globally valid. Because different identifiers may belong to the same person, the development and evaluation data are screened for cross-identifier duplication using a procedure independent of HAPPEN's learned matcher. Pairwise matching is treated as an intermediate result: candidate links undergo human review, and reviewer-supported links are reconciled into subject-level groups. The complete audit was evaluated at national scale across 95,129 scans from 21 datasets, with sensitivity independently assessed against 54 duplicate pairs confirmed by genome-wide genetic data. The same workflow was then run by an independent institution without retraining or transferring imaging data. Together, these elements extend image-based subject matching into a deployable methodology for repository identity auditing.

## 3. METHODS

HAPPEN contains two detection stages, with human review applied to Stage II candidates (Figure 2). The workflow operates on a repository snapshot comprising the scans selected for an audit and their associated dataset, recorded subject, and session identifiers. Evidence of duplication is derived only from image content, and the identifiers instead serve as provenance metadata for filtering matches and stratifying the reported results.

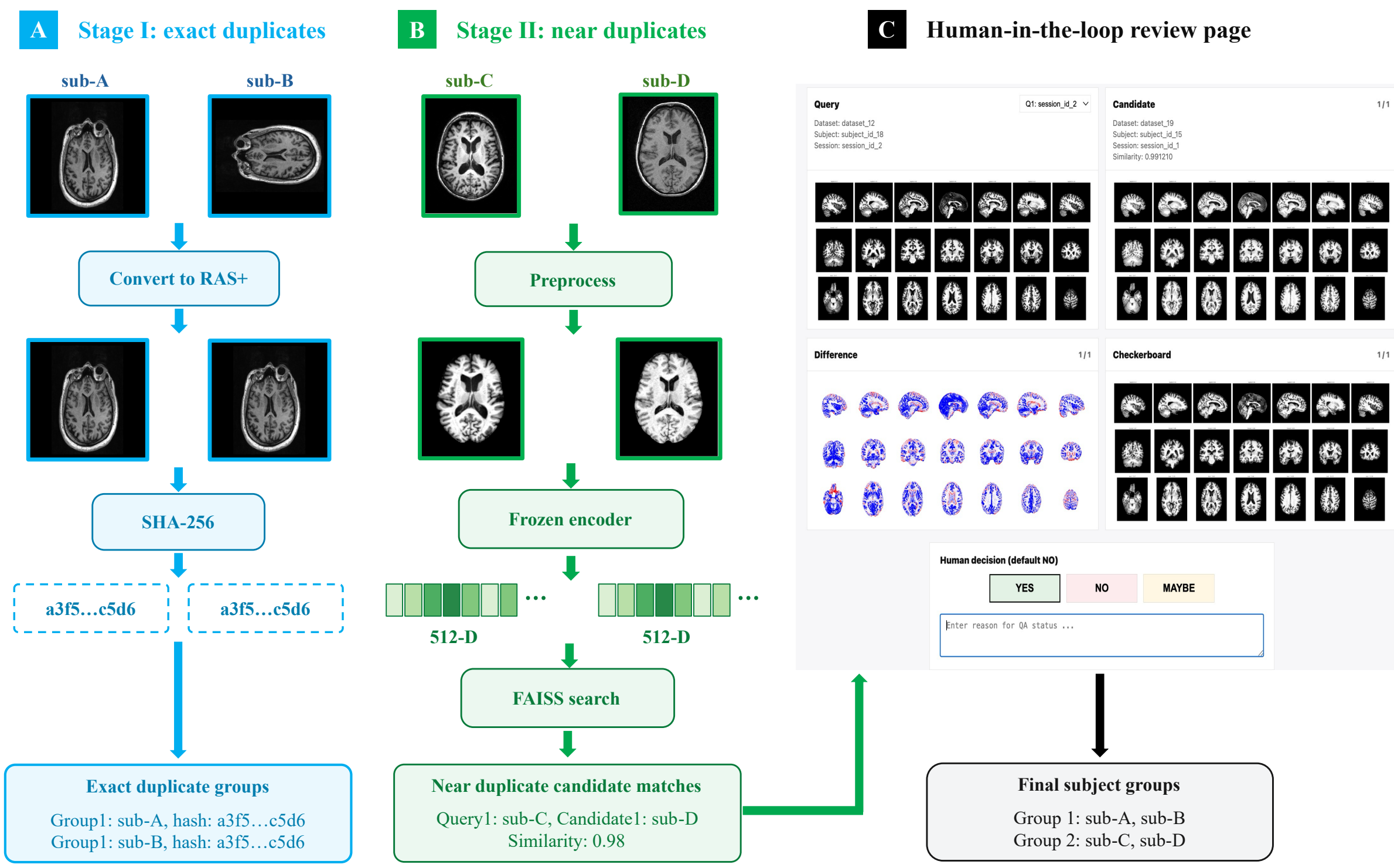

**Figure 2. Overview of the HAPPEN workflow.** (A) Stage I detects exact duplicates from image content. Each scan is reoriented to canonical RAS+ orientation, and a SHA-256 fingerprint is computed from the image voxel bytes. Scans with identical fingerprints are assigned to the same exact-duplicate group. (B) Stage II generates near-duplicate candidates. After preprocessing, each scan is mapped by the encoder to a 512-dimensional embedding. The embeddings are indexed using FAISS, and the nearest neighbors of each query are retrieved by cosine similarity. (C) Candidate pairs are evaluated in the locally hosted review interface, which presents sagittal, coronal, and axial views of the query and candidate scans, together with difference and checkerboard views, cosine similarity, and provenance metadata. Reviewers record *Yes*, *No*, or *Maybe* decisions and may provide an optional reason. Accepted near-duplicate pairs are combined with the Stage I exact-duplicate groups to form the final subject groups.

### 3.1 Stage I: exact-duplicate detection

**Scope.** Stage I defines exact duplicates according to image content rather than file representation. Two scans are considered exact duplicates when their voxel arrays are identical after canonical reorientation to Right-Anterior-Superior (RAS+) orientation. This operation may reorder or flip voxel axes but does not interpolate the image. The definition therefore naturally covers two common cases: byte-identical files; and files that preserve the same voxel content despite differences in compression, header metadata, or stored orientation. Images with different voxel values, including differences introduced by quantization, interpolation, or another intensity-altering operation, fall outside Stage I and may instead be identified in Stage II.

**Hash fingerprinting.** To detect exact duplication at repository scale, we use SHA-256 based hash grouping [25], [26]. Under our definition above, each scan is loaded using NiBabel, with NIfTI intensity scaling applied during voxel-array access and converted to RAS+ orientation [27]. The resulting canonical voxel array is converted to little-endian float32 and serialized in C order. A SHA-256 fingerprint is computed from the resulting bytes without incorporating header fields or other non-image metadata [25], [26]. Scans sharing the same fingerprint are placed in the same exact-duplicate group (Figure 2A).

### 3.2 Stage II: near-duplicate detection and human-in-the-loop review

**Scope.** Stage II operates on the repository snapshot after Stage I. One scan from each exact-duplicate group is retained, while redundant copies are excluded from subsequent retrieval. Stage II searches the remaining scans for non-identical pairs that may originate from the same individual but are assigned to different repository subject records. A repository subject record is defined by the combination of a dataset identifier and a recorded subject identifier. Unlike the voxel equality used in Stage I, embedding similarity indicates that two scans may originate from the same individual but does not establish that relationship. Stage II therefore returns scan-level candidate pairs rather than final decisions.

**Approach.** HAPPEN generates these candidates in two steps. First, HAPPEN represents each scan using a learned identity-embedding model so that scans from the same real individual remain close in the embedding space. It then applies approximate nearest-neighbor (ANN) retrieval to search this space efficiently at repository scale.

**Preprocessing.** Before a scan is passed to the model, HAPPEN applies a fixed preprocessing pipeline: (1) N4 bias field correction [28]; (2) skull stripping using DeepBet [29]; (3) rigid registration to the 1-mm isotropic ICBM 152 nonlinear symmetric 2009c template using ANTs, with the corresponding brain mask transformed using nearest-neighbor interpolation [30], [31], [32], [33]; (4) a second N4 bias field correction after alignment to reduce residual intensity bias; (5) min-max intensity normalization within the brain mask; (6) masking of voxels outside the brain; and (7) center-cropping and extraction of 32 central axial slices. These steps yield a fixed model input of $160 \times 192 \times 32$ voxels.

**Identity-embedding model.** The model is trained using supervised contrastive learning [34]. Scans from the same repository subject record and their augmented versions are treated as positive views, whereas views derived from different repository subject records are treated as negative views. This objective encourages the model to preserve subject-specific information shared across longitudinal scans while reducing sensitivity to acquisition-related variation. A 3D ResNet-like encoder with squeeze-and-excitation modules maps each input to a 512-dimensional representation [35]. The resulting representation is L2-normalized before retrieval.

**Candidate generation.** Only scans that successfully complete preprocessing and embedding computation enter candidate generation. Failures are recorded without interrupting the remaining audit. To avoid exhaustive pairwise image comparison, HAPPEN builds a Facebook AI Similarity Search (FAISS) ANN index from the scan embeddings [36]. For each query scan, the 500 nearest neighbors are retrieved according to cosine similarity. After retrieval, matches within the same repository subject record are excluded as expected longitudinal observations; only cross-record matches above the prespecified operating threshold are retained as candidates. The retained matches are consolidated into unique scan-level candidate pairs and passed to human review (Figure 2B).

**Human-in-the-loop review.** HAPPEN provides a locally hosted web interface for reviewing each query-neighbor pair generated by Stage II. On one page, it displays sagittal, coronal, and axial views of both scans, a difference map, a checkerboard view, cosine similarity, and scan provenance. Review images are preloaded for rapid navigation between pairs (Figure 2C). The reviewer records the decision on the same page using *Yes* (same individual), *No* (different individuals), or *Maybe* (uncertain) buttons. *Maybe* allows an uncertain pair to be retained without forcing a *Yes* or *No* decision. The reviewer can also enter the reason for the decision in a free-text box. Each decision and its optional reason are automatically saved in a scan-level CSV file. Existing decisions are reloaded when the review interface is restarted, allowing the review to be paused and resumed. This human-in-the-loop framework was adapted from design details from Kim et al. [37]. The resulting decision file is used to construct the repository-level audit outputs described in Section 3.3.

### 3.3 Duplicate reporting

**Grouping.** HAPPEN reports Stage I findings as scan-level exact-duplicate groups and Stage II findings as subject-level near-duplicate groups. Stage I places scans sharing the same voxel-array fingerprint in the same exact-duplicate group. Stage II constructs subject-level groups from the reviewed query-neighbor pairs. Repository subject records linked by at least one *Yes* pair are then treated as connected. Records connected either directly or through another record are placed in the same group. *No* and *Maybe* pairs remain in the scan-level review decision file but do not contribute to subject-level grouping. For final subject-level reporting, repository subject records linked by either Stage I or Stage II were consolidated. Records connected directly or through another record were placed in the same final HAPPEN subject group.

**Record-level reporting.** After grouping, repository identifiers are used to stratify the findings by provenance. Exact-duplicate findings are reported at four provenance levels. Within-session findings share the same dataset, recorded subject, and session identifiers. Within-subject findings span sessions under the same dataset and recorded subject identifier. Within-dataset findings span different recorded subject identifiers within one dataset, whereas across-dataset findings span multiple datasets. Near-duplicate findings are reported only at the within-dataset and across-dataset levels. A group containing more than two scans or subject records may include relationships at more than one provenance level. These levels therefore describe the relationships within a group rather than mutually exclusive labels assigned to the entire group.

**Repository-level reporting.** While the record-level reporting describes individual findings, HAPPEN uses interactive duplication matrices to summarize how duplication is distributed across the repository. Separate matrices are generated from the scan-level exact-duplicate groups and the subject-level near-duplicate groups. These matrices provide an overall view of whether duplication is concentrated within individual datasets or extends across datasets.

### 3.4 Implementation and reproducibility

**Incremental auditing.** HAPPEN saves completed computations and reuses them in later runs. In Stage I, hashes are reused for paths that have already been processed within the same repository snapshot. In Stage II, each scan is assigned a SHA-256 identifier based on its file content. If the file is unchanged, its previously computed embedding is reused. Only new or modified scans require preprocessing and embedding inference. Interrupted runs can also continue from completed work. Candidate retrieval and reporting can then be updated for the current repository snapshot. This design supports repeated audits as a repository grows or releases new versions.

**Portable execution.** HAPPEN supports execution via Docker and Apptainer and is released as a versioned container image. Users control both the auditing pipeline and the review interface through a TOML configuration file. A copy of the configuration is saved with each audit. All stochastic components use fixed random seeds.

## 4. EXPERIMENTAL DESIGN

### 4.1 Data sources

**Study repository.** This study was developed using an aggregated MRI repository that includes the datasets in the 2025 diffusion weighted imaging release of the Alzheimer's Disease Sequencing Project Phenotype Harmonization Consortium (ADSP-PHC, U24-AG074855) [38], together with additional datasets. At the time of analysis, the repository comprised 21 datasets in total, including 40,218 recorded subjects and 95,129 T1-weighted brain MRI scans. Within the repository, the Baltimore Longitudinal Study of Aging (BLSA) [39] and Vanderbilt Memory and Aging Project (VMAP) [40] were set aside as withheld evaluation datasets, while the remaining 19 datasets provided the source population for model development and internal evaluation. BLSA and VMAP did not inform model fitting, model selection, or operating-threshold selection. The full list of contributing datasets is provided in Supplementary table S1.

Data used in the preparation of this article were obtained from the Alzheimer's Disease Neuroimaging Initiative (ADNI) database (adni.loni.usc.edu). The ADNI was launched in 2003 as a public-private partnership, led by Principal Investigator Michael W. Weiner, MD. The primary goal of ADNI has been to test whether serial magnetic resonance imaging (MRI), positron emission tomography (PET), other biological markers, and clinical and neuropsychological assessment can be combined to measure the progression of mild cognitive impairment (MCI) and early Alzheimer's disease (AD).

**UPenn repository.** In addition to the study repository, we conducted an external-site deployment at the University of Pennsylvania (UPenn) using an independent repository containing 6 datasets, 14,690 recorded subjects and 22,386 T1-weighted brain MRI scans. Dataset inclusion was determined by the UPenn team, and no UPenn data were used to train or select the Stage II model. The deployment protocol is described in Section 4.4.

### 4.2 Duplicate-aware construction of the development and test datasets

Because Stage II is designed to detect identity duplication, duplicates in its own development or evaluation data may inflate its apparent performance. We therefore constructed the development and test datasets to meet two requirements: Stage II needed repeated scans from the same recorded subject for supervised contrastive learning, and identity duplication needed to be controlled before model training and evaluation.

**Initial repository screening.** Before selecting the development and test datasets, we conducted an initial repository screening. Using Stage I (Section 3.1), we consolidated each exact-duplicate group by retaining one representative scan and excluding redundant copies from subsequent dataset construction. We then applied the scan-level quality assessment method developed by Kim et al. [37] to the remaining original images and excluded corrupted or otherwise unusable scans. The remaining scans formed the eligible repository used for dataset construction (Figure 3).

**Model-development dataset.** Supervised contrastive learning requires multiple non-identical scans from the same recorded subject to construct positive views. Given the computational cost of training on the full pool of eligible subjects, we constructed a smaller model-development subset from the 19 available datasets. The primary pool comprised 5,000 subjects with at least three non-identical scans. To preserve representation from smaller datasets, all eligible subjects were included from datasets containing no more than 100 such subjects; the remaining positions were filled by randomly sampling eligible subjects from datasets with larger pools. We then added 457 subjects with exactly two non-identical scans so that every development dataset contributed at least 100 subjects. Three scans were retained for each subject in the primary pool, and both scans were retained for each additional subject. When session labels were available, scans from different sessions were preferentially selected to increase longitudinal variation among positive views.

**Withheld datasets.** BLSA and VMAP served as withheld test datasets. Because evaluating identity consistency requires more than one scan from each subject, subjects with only one available scan were excluded.

**Preprocessing and quality assessment.** Then the same fixed preprocessing pipeline described in Section 3.2 was applied to the model-development dataset and the two withheld test sets. Although the initial quality assessment excluded unusable original images, additional failures could arise during preprocessing, such as unsuccessful skull stripping or registration. We therefore applied the same scan-level quality assessment method to the preprocessed outputs and excluded scans that failed this second assessment before near-duplicate screening.

**PCA near-duplicate screening.** We next screened the model-development dataset and the two withheld test sets for pre-existing near duplicates. To avoid circularity, Stage II was not used for this step, because doing so would make dataset construction depend on the model being trained and evaluated. Instead, we used unsupervised PCA-based screening followed by visual review. After preprocessing and the second quality assessment, the model-development dataset, BLSA, and VMAP were pooled to fit an Incremental PCA model using the settings described in Supplementary Section S2.1. Each scan was projected into the resulting PCA space, and scan pairs from different recorded subjects were treated as possible near duplicates when their cosine similarity was high. One researcher visually reviewed these candidates, and scans judged to be near duplicates were removed. This procedure was used only for dataset cleaning and was separate from the PCA retrieval baseline evaluated later, which was fitted only on the training split.

**Data partitioning.** To prevent scans from the same recorded subject from appearing in different stages of model development, the cleaned model-development dataset was divided at the subject level. Within each contributing dataset, subjects were assigned to the training, validation, and internal test splits in an 8:1:1 ratio, with all scans from a given subject kept in the same split. The training split was used to fit Stage II, and the validation split was used for model selection and operating-threshold selection. The internal test split was reserved for evaluation, together with the separately withheld BLSA and VMAP datasets. Final dataset sizes and their experimental roles are reported in Table 1.

**Training views.** After the subject-level split, training views were constructed only from the training split. To ensure that subjects represented by two and three scans contributed the same number of views to supervised contrastive learning, we constructed six views for each training subject. For subjects with three scans, one augmented view was generated from each scan, producing three original and three augmented views. For subjects with two scans, two augmented views were generated from each scan, producing two original and four augmented views. Augmentation was performed using TorchIO [41] and was designed to introduce intensity variation and common MRI artifacts. Validation and test scans were not augmented. The transformations and their probabilities are provided in Supplementary table S2.

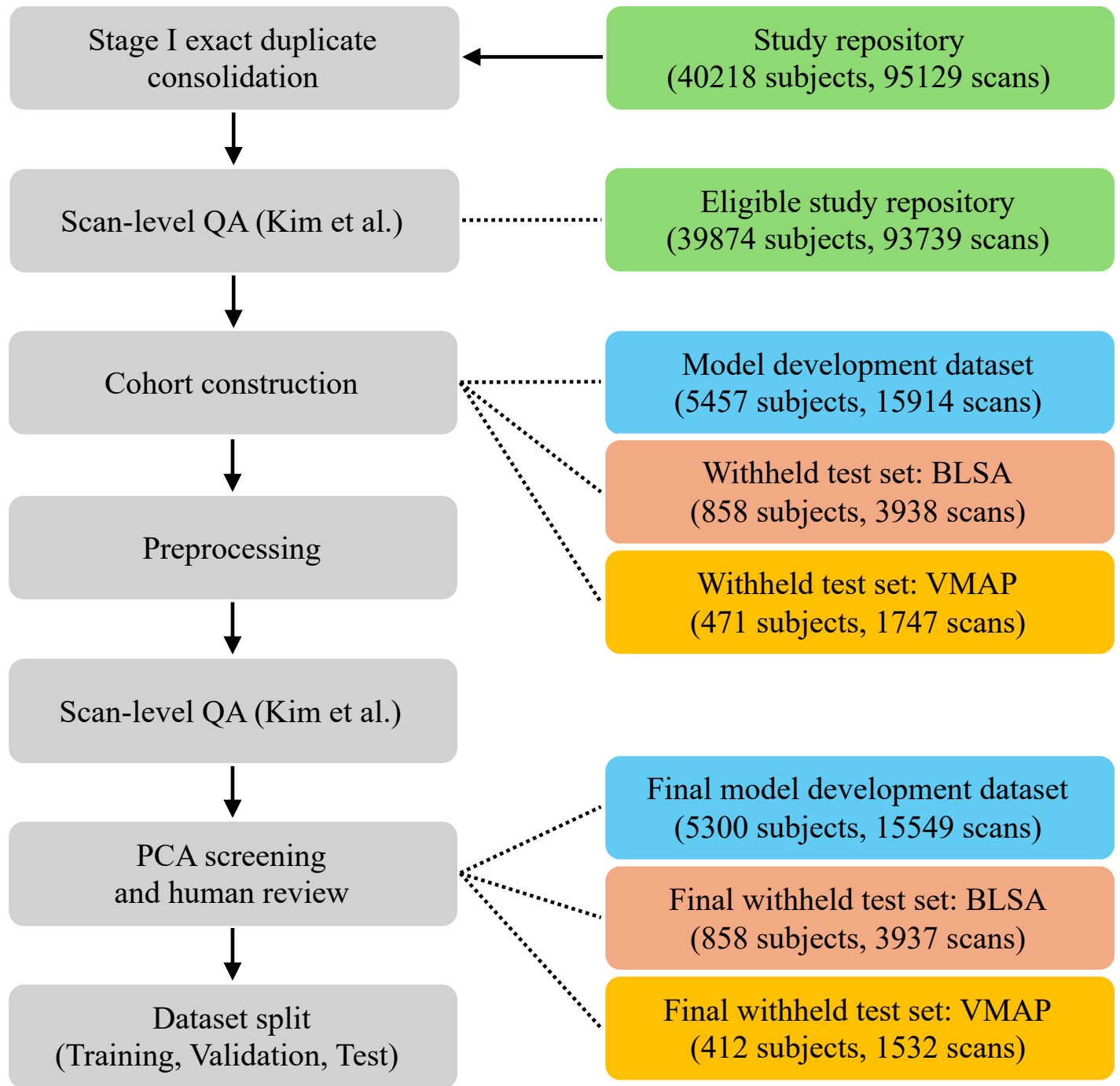

**Figure 3. Duplicate-aware dataset construction.** The entire construction workflow was designed to prevent identity duplication from biasing model development or evaluation. After exact-duplicate consolidation and initial scan-level quality assessment, the model-development dataset was sampled and BLSA and VMAP were reserved for withheld testing; only these three datasets entered the remaining workflow. Following preprocessing, the same quality assessment was repeated to remove failed outputs, and the three datasets were jointly screened for near duplicates using PCA-based screening and human review. The final model-development dataset was then split by subject into training, validation, and internal test sets, while BLSA and VMAP remained withheld.

**Table 1. Final datasets used for Stage II model development and HAPPEN evaluation.** Study repository counts represent the repository before duplicate removal and quality assessment. Counts for the model-development dataset, BLSA, and VMAP are reported after preprocessing, post-preprocessing quality assessment, and near-duplicate screening. UPenn repository counts describe the independent repository used for external-site deployment.

| Data source | Experimental role | Datasets | Subjects | Scans |
|---|---|---|---|---|
| Study repository | Stage II development and end-to-end deployment | 21 | 40,218 | 95,129 |
| | Stage II training set | 19 | 4,242 | 12,444 |
| Model-development dataset | Stage II validation set (model and threshold selection) | 19 | 529 | 1,554 |
| | Stage II internal test set | 19 | 529 | 1,551 |
| BLSA | Withheld test set | 1 | 858 | 3,937 |
| VMAP | Withheld test set | 1 | 412 | 1,532 |
| UPenn repository | External-site deployment | 6 | 14,690 | 22,386 |

### 4.3 Evaluation of Stage I and Stage II

**Stage I evaluation.** Stage I was evaluated using controlled variants generated from 100 randomly selected T1-weighted MRI scans. For each source scan, we created three positive controls that preserved its canonical voxel content: a byte-identical copy, a copy with modified metadata but unchanged voxel values, and an orientation-modified copy produced by flipping voxel axes and updating the affine without interpolation. We also created one negative control by changing a single voxel value. Stage I was expected to assign the source scan and its three positive controls to the same exact-duplicate group while assigning the one-voxel-modified control a different fingerprint.

**Stage II evaluation.** We evaluated Stage II in two steps. The retrieval evaluation first tested whether the learned embedding captured subject identity information. Then the operating-threshold selection determined the default similarity cutoff used by HAPPEN to decide which retrieved scan pairs should be retained as candidates for human review.

**Retrieval evaluation.** The frozen Stage II encoder was evaluated on the internal test split, BLSA, and VMAP. Within each test dataset, every scan was used as a query and ranked against all remaining scans according to cosine similarity. Scans from the same repository subject record were treated as positives, and scans from different records were treated as negatives. Performance was measured using the all-positive rate (APR), mean average precision (mAP), and ROC-AUC. APR was the proportion of queries for which every positive scan ranked above every negative scan; mAP summarized the positions of positive scans throughout the ranked results; and ROC-AUC measured the overall separation between same-subject and different-subject pairs. As baselines, a separate Incremental PCA model was fitted using only the training split, and the 3D SIFT-Rank and soft-Jaccard implementation was evaluated under two input configurations: the FreeSurfer skull-stripped input described by Chauvin et al. (3D-SIFT-FS) and the same preprocessing method used by HAPPEN (3D-SIFT-Shared) [24]. We also evaluated the computational efficiency on 200 scans using 32 workers. We report median representation time with its interquartile range (IQR), mean search time per query, and median representation size. Training and preprocessing time were excluded. PCA and HAPPEN used flat FAISS search, whereas 3D SIFT-Rank used its official search method.

**Operating-threshold selection.** The operating threshold controls a practical trade-off: a lower threshold can retain more potential near-duplicate matches but also produces more candidate pairs for human review, whereas a higher threshold reduces the review set at the risk of missing matches. Because ground-truth identity labels across different repository subject records were not available at large scale, we approximated this trade-off in the validation split using longitudinal pairs from the same repository subject record as known same-subject references. For this analysis, let $S$ denote all such same-subject pairs and let $C_\lambda$ denote the simulated review set formed by all pairs retained at threshold $\lambda$. We calculated a **recall proxy**,

$$R(\lambda) = \frac{|S \cap C_\lambda|}{|S|}, \quad (1)$$

which measured the proportion of same-subject pairs retained, and a **review-efficiency proxy**,

$$E(\lambda) = \frac{|S \cap C_\lambda|}{|C_\lambda|}, \quad (2)$$

which measured the concentration of same-subject pairs in the simulated review set. We compared these measures across 12 thresholds ranging from 0.85 to 0.96 and selected a practical default based on their trade-off. The selected threshold was subsequently used for deployment in the study repository.

### 4.4 End-to-end pipeline evaluation

The evaluations in Section 4.3 assessed the behavior of Stage I and Stage II, whereas the end-to-end deployments examined computational feasibility, human-review requirements, and use in independently governed repository settings.

**Deployment in the study repository.** We applied the complete HAPPEN workflow described in Section 3 to the original study repository summarized in Table 1, using the frozen Stage II encoder and the default operating threshold selected in Section 4.3. The pipeline was executed on a workstation with 64 CPU cores, 512 GB of host memory, and an NVIDIA RTX A6000 GPU with 48 GB of memory, with 32 parallel processes. One researcher reviewed all Stage II candidate pairs using the interface described in Section 3.2. Deployment performance was assessed using wall time, cumulative CPU time, GPU inference time, and total human-review time. Pipeline findings were reported separately for the two stages: Stage I results included the exact-duplicate groups and scans involved, whereas Stage II results included the candidate pairs, review decisions, and resulting subject-level near-duplicate groups. For both stages, we also examined whether the identified duplication occurred within or across datasets.

**Genetic-reference sensitivity evaluation.** Following deployment in the study repository and human review, we compared the final HAPPEN subject groups with 54 duplication pairs identified using genome-wide genotype data (PI_HAT > 0.9). These pairs linked subject records from different datasets. None of the reference subjects was included in the Stage II training split. For each reference pair, we determined whether HAPPEN placed both records in the same final subject group. Sensitivity was calculated as the proportion of reference pairs recovered and was reported with a 95% Wilson confidence interval.

**External-site repository deployment.** The UPenn deployment was designed to assess whether HAPPEN could be transferred to and operated by an independent institution with minimal involvement from the model-development team. The handoff consisted of the same versioned Apptainer image used in this study together with run-configuration instructions. UPenn independently selected the datasets, executed the complete workflow within its local computing environment, and assigned one local reviewer to assess the Stage II candidates. The review criteria were not harmonized with those used in the study repository deployment. The external reviewer deliberately applied a more conservative standard; therefore, the two deployments were not designed as a comparison of model precision. The frozen Stage II encoder was used without retraining, while UPenn independently set the operating threshold to 0.99 according to its local review budget. The job was allocated 32 CPU cores and one NVIDIA RTX A6000 GPU and was run using 32 parallel processes. All MRI data remained within the UPenn environment throughout the deployment.

## 5. RESULTS

### 5.1 Stage I evaluation

Stage I behaved exactly as specified in the controlled evaluation. All 300 positive controls shared the fingerprint of their corresponding source scan, whereas all 100 one-voxel-modified controls received a different fingerprint. Stage I was therefore invariant to changes in file metadata and stored orientation when canonical voxel content was preserved, but sensitive to changes in voxel values.

### 5.2 Stage II evaluation

**Table 2. Retrieval performance and computational efficiency of HAPPEN and the baseline methods.** Panel A reports APR, mAP, and ROC-AUC with 95% confidence intervals for the three retrieval cohorts. Panel B reports median representation time (IQR), mean search time per query, and median representation size from the controlled 200-scan benchmark.

**Panel A. Retrieval accuracy**

| Cohort | Method | APR | mAP | ROC-AUC |
|---|---|---|---|---|
| Internal testing split | PCA | 0.9378 [0.9177, 0.9565] | 0.9667 [0.9553, 0.9781] | 0.9949 [0.9907, 0.9978] |
| | 3D-SIFT-FS | 0.9437 [0.9206, 0.9655] | 0.9689 [0.9543, 0.9820] | 0.9952 [0.9913, 0.9984] |
| | 3D-SIFT-Shared | 0.9845 [0.9750, 0.9929] | 0.9941 [0.9896, 0.9976] | 0.9998 [0.9995, 1.0000] |
| | HAPPEN | 0.9956 [0.9898, 1.0000] | 0.9975 [0.9934, 1.0000] | 0.9999 [0.9998, 1.0000] |
| BLSA | PCA | 0.7917 [0.7509, 0.8348] | 0.9256 [0.9055, 0.9431] | 0.9862 [0.9820, 0.9902] |
| | 3D-SIFT-FS | 0.7606 [0.7151, 0.8082] | 0.9018 [0.8817, 0.9226] | 0.9367 [0.9228, 0.9508] |
| | 3D-SIFT-Shared | 0.9881 [0.9799, 0.9950] | 0.9997 [0.9995, 0.9999] | 1.0000 [1.0000, 1.0000] |
| | HAPPEN | 0.9975 [0.9933, 0.9997] | 0.9997 [0.9994, 1.0000] | 1.0000 [1.0000, 1.0000] |
| VMAP | PCA | 0.9639 [0.9455, 0.9804] | 0.9918 [0.9857, 0.9969] | 0.9996 [0.9990, 1.0000] |
| | 3D-SIFT-FS | 1.0000 [1.0000, 1.0000] | 1.0000 [1.0000, 1.0000] | 1.0000 [1.0000, 1.0000] |
| | 3D-SIFT-Shared | 1.0000 [1.0000, 1.0000] | 1.0000 [1.0000, 1.0000] | 1.0000 [1.0000, 1.0000] |
| | HAPPEN | 1.0000 [1.0000, 1.0000] | 1.0000 [1.0000, 1.0000] | 1.0000 [1.0000, 1.0000] |

**Panel B. Computational efficiency**

| Method | Representation time, s / scan | Search time, ms / query | Size, KiB / scan |
|---|---|---|---|
| PCA | 10.52 (5.55) | 0.013 | 0.69 |
| 3D-SIFT-FS | 5.12 (0.63) | 37.56 | 971.65 |
| 3D-SIFT-Shared | 1.89 (0.33) | 35.32 | 953.47 |
| HAPPEN | 0.51 (0.14) | 0.018 | 2.00 |

**Retrieval evaluation.** The HAPPEN Stage II model outperformed the PCA baseline on all three metrics across all evaluation cohorts (Table 2). Compared with 3D SIFT-Rank, HAPPEN achieved the highest APR in the internal test and BLSA cohorts and matched the ceiling performance observed on VMAP. SIFT performance was strongly influenced by preprocessing: the shared preprocessing configuration markedly outperformed the FreeSurfer-based configuration and narrowed the difference from HAPPEN, particularly for mAP and ROC-AUC. In the computational efficiency test, HAPPEN produced a substantially smaller representation size and supported faster representation generation and retrieval than SIFT (Table 2B).

**Operating-threshold selection.** Figure 4 shows the trade-off between retaining more known same-subject pairs and maintaining their concentration in the simulated review set. Lowering the threshold increased the recall proxy but reduced the review-efficiency proxy. We selected $\lambda = 0.92$ because it retained more than 98% of known same-subject pairs while avoiding the sharper loss of review efficiency observed at lower thresholds. This threshold was used as the default for the study repository deployment and represents a practical operating point rather than a universally optimal point.

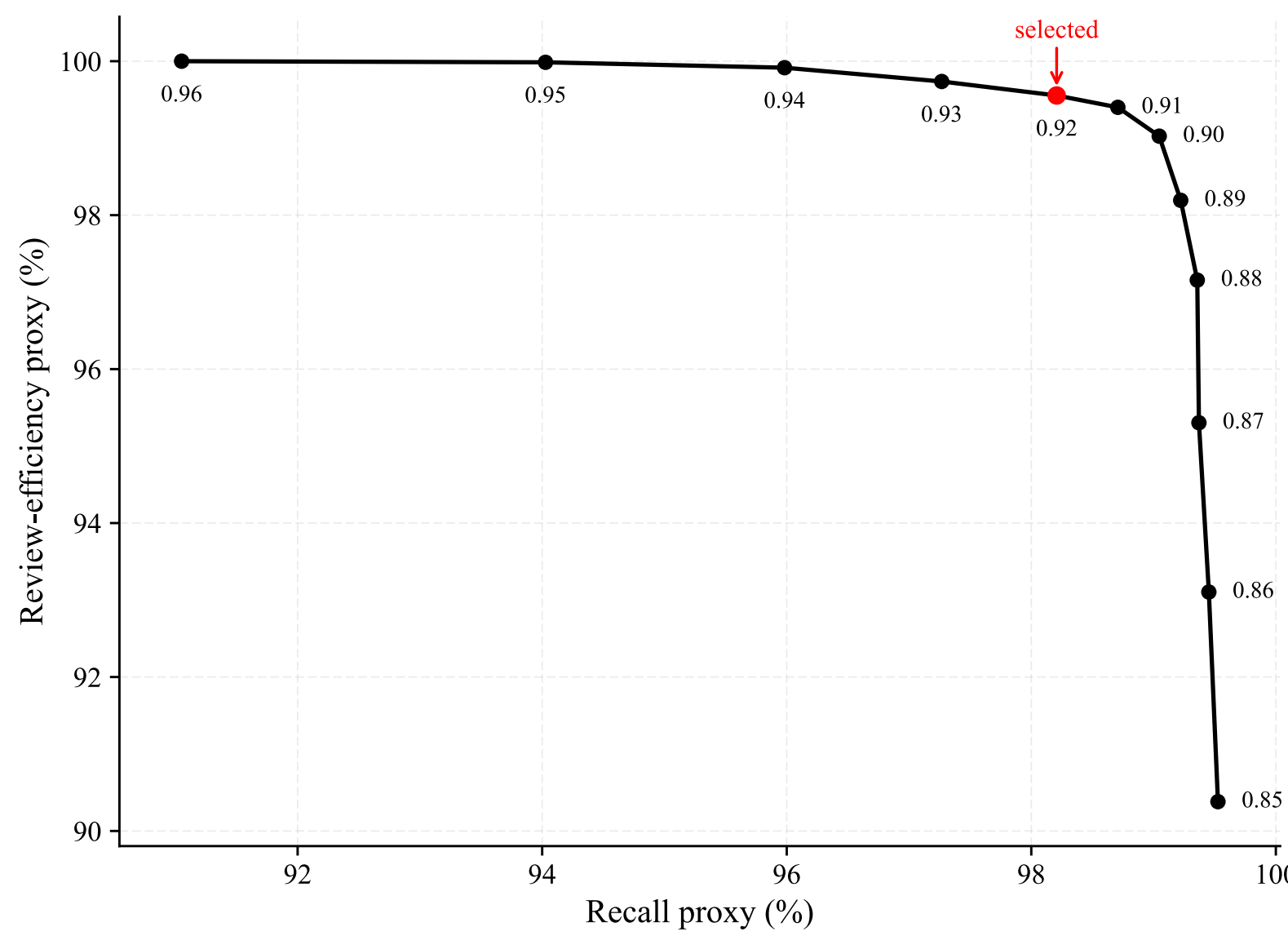

**Figure 4. Operating-threshold selection.** The x-axis shows the recall proxy, and the y-axis shows the review-efficiency proxy. Each point represents one similarity threshold; the selected default threshold ($\lambda = 0.92$) is highlighted in red.

### 5.3 End-to-end pipeline evaluation

**Table 3. End-to-end deployment results in two independent repositories.** Panel A reports audit outcomes from the two deployments, including Stage I results, Stage II candidate generation, and review results. Panel B reports deployment performance. Wall time denotes elapsed time, CPU time denotes cumulative time across worker processes, and GPU time denotes model-inference time. Normalized values are reported as hours per 100,000 input scans. Review criteria were independently determined and were not harmonized; the external reviewer deliberately applied a more conservative standard. Therefore, review outcomes are site-specific and are not directly comparable measures of model precision.

**Panel A. Audit and review outcomes**

| Stage | Measure | Study repository | External-site repository |
|---|---|---|---|
| Repository | Recorded subject identifiers | 40,218 | 14,690 |
| | Scans | 95,129 | 22,386 |
| Stage I | Exact-duplicate groups | 1,316 | 10 |
| | Scans in exact-duplicate groups | 2,636 | 20 |
| Stage II pipeline | Candidate scan pairs | 9,802 | 3,323 |
| Stage II local review | Candidate pairs judged likely near duplicate | 5,123 | 303 |
| | Uncertain / not accepted | 221 / 4,458 | 1 / 3,019 |
| Stage II grouping | Subject-level groups | 1,275 | 255 |
| | Recorded identifiers in subject-level groups | 2,570 | 535 |

**Panel B. Computational efficiency**

| Measure | Unit | Study repository | External-site repository |
|---|---|---|---|
| Repository size | scans | 95,129 | 22,386 |
| Total wall time | hours | 164.90 | 80.49 |
| Normalized wall time | hours per 100k scans | 173.34 | 359.57 |
| Normalized CPU time | hours per 100k scans | 8371.41 | 11381.35 |
| Normalized GPU time | hours per 100k scans | 0.02 | 0.08 |
| Total review time | hours | 4.5 | 6 |

### 5.3.1 Deployment in the study repository

In the study repository, Stage I identified 1,316 exact-duplicate groups involving 2.8% of all scans; 55.8% of these groups spanned datasets (Table 3). Following exact-duplicate consolidation, 93,809 scans entered Stage II; 58 (0.06%) failed automated preprocessing and were excluded from embedding and candidate generation. After human review, approximately half of the Stage II candidate pairs were judged likely to be near duplicates. These pairs formed 1,275 subject-level groups involving 6.4% of all recorded identifiers; most of these groups (82.4%) linked records from different datasets.

**Genetic-reference sensitivity evaluation.** HAPPEN recovered all 54 genetic-reference pairs available for evaluation, corresponding to a sensitivity of 100% (95% CI, 93.4%-100%).

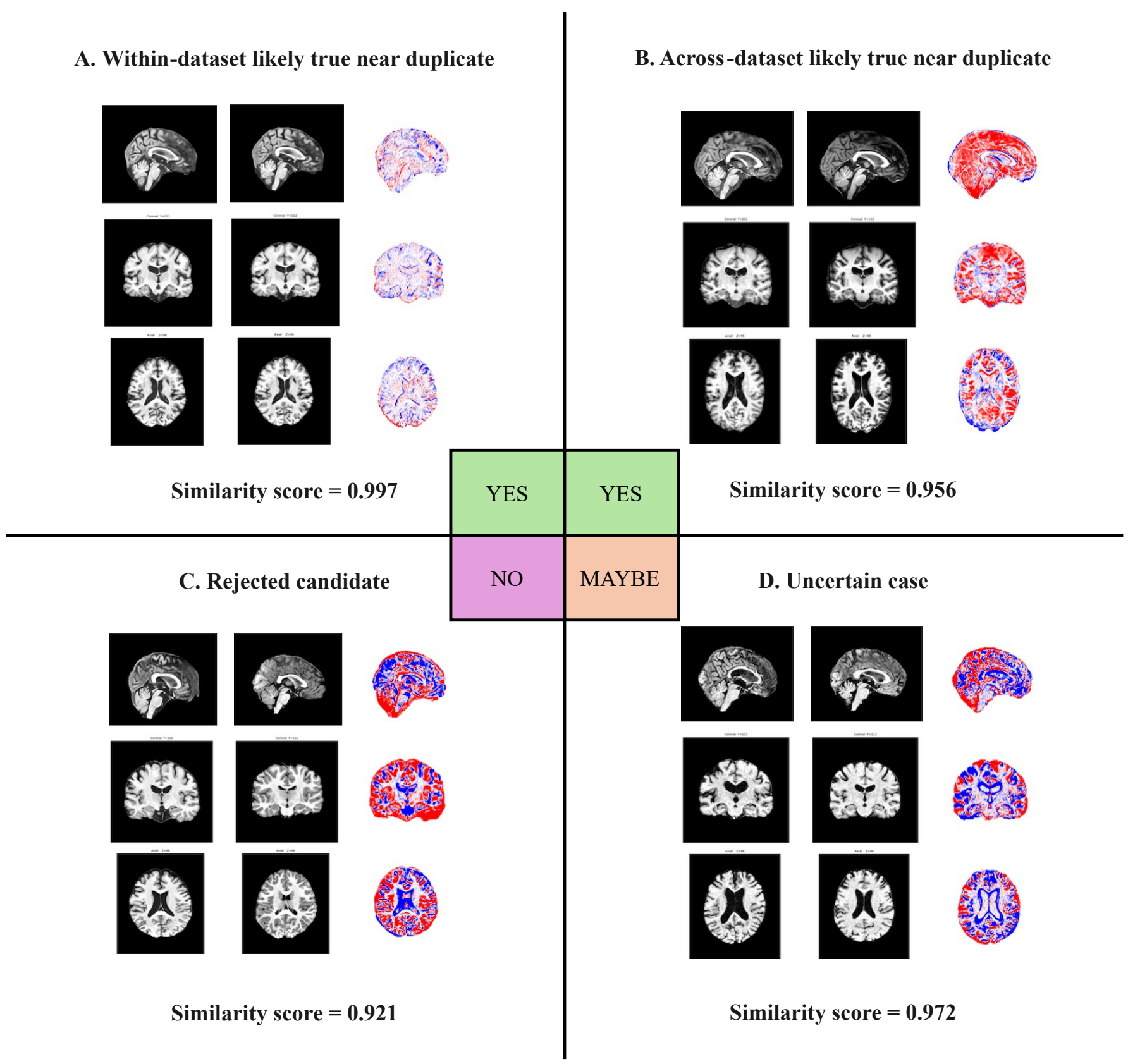

**Figure 5. Qualitative examples from human-in-the-loop review.** Panels A and B show within- and across-dataset near-duplicate pairs, respectively; Panel C shows a pair not accepted as a near duplicate, and Panel D shows an uncertain pair. Each panel presents the query scan, candidate scan, and query-minus-candidate difference image. Similarity scores are shown below each panel. All scans are shown after preprocessing.

**Qualitative examples.** Figure 5 illustrates that the similarity score alone did not determine the review outcome. The uncertain pair in Panel D received a higher similarity score (0.972) than the reviewer-supported across-dataset pair in Panel B (0.956). Final decisions therefore also required consideration of scan provenance and anatomical correspondence.

### 5.3.2 External-site repository deployment

In the external-site repository, exact duplication was uncommon, involving fewer than 0.1% of scans (Table 3). Under the site's independently determined, more conservative review standard, 9.1% of the Stage II candidate pairs were judged likely to be near duplicates; this fraction is not directly comparable with that of the study repository deployment. Nearly all resulting subject-level groups (97.6%) linked records from different datasets.

### 5.3.3 Deployment performance

Deployment in the study repository was completed in under seven days. In both settings, GPU inference time remained small relative to cumulative CPU time, indicating that computational cost was dominated by CPU-side operations. The two sites used different computing environments, so their runtimes are not directly comparable.

## 6. DISCUSSION

Controlled evaluations supported the reliability of both stages of HAPPEN. Stage I behaved as expected in every controlled test. Stage II exceeded PCA and 3D-SIFT-FS in the internal and BLSA cohorts and achieved comparable or higher retrieval point estimates than 3D-SIFT-Shared. The computational efficiency test further showed that HAPPEN's fixed-length embedding was substantially faster to generate and search, and much smaller than the variable-length SIFT keypoint fingerprint. Importantly, Stage II achieved near-ceiling performance not only in the internal test split but also in the two withheld cohorts. This consistency supports applying the frozen encoder to datasets that were not used in model development. In addition to the retrieval results, the genetic-reference evaluation showed that the complete workflow can recover duplication cases identified by genome-wide genotype data.

The study shows that identity duplication represents a structural risk in aggregated MRI repositories. In the study repository, exact-duplicate groups involved 2.8% of all scans, while reviewer-supported near-duplicate groups involved 6.4% of all recorded identifiers. Duplicate groups were identified both within individual datasets and across dataset boundaries. Of the near-duplicate groups, 82% involved more than one dataset. Cross-dataset links were also identified in the independent UPenn deployment. These findings are consistent with the aggregation, re-keying, and data-transfer pathways described in the Introduction. They indicate that subject identifiers created within individual datasets cannot be assumed to remain unique when datasets, versions, and releases are combined. Identity duplication should therefore be treated as a quality assurance concern whenever datasets or releases are combined.

The external-site deployment at UPenn demonstrated that HAPPEN could be operated end to end outside its development environment. No imaging data were shared between the development and UPenn teams, and UPenn researchers independently completed the full workflow using the containerized pipeline and frozen encoder.

Importantly, the operating decisions remained local. UPenn researchers chose a higher similarity threshold and applied a site-specific review policy. Rather than imposing a universal operating point, this flexibility allows each repository to balance sensitivity, candidate volume, and review effort according to its own priorities and resources. This successful deployment at UPenn also indicates that the need for identity-level repository checks extends beyond the environment in which HAPPEN was developed.

The applications of HAPPEN span both repository management and the downstream use of aggregated imaging data. Within the repository lifecycle, HAPPEN can be run after new images are ingested, when dataset versions or releases are combined, and before a new release is distributed. Confirmed links can prompt repository teams to compare provenance records, determine why the same scan or participant appears under different identifiers, and document the relationship in the release. At the analysis stage, all scans linked to the same individual can be assigned a common group label. This allows researchers to count independent participants correctly and keep each individual within a single training, validation, or test set. Model performance can then be evaluated on individuals who were not represented during training. HAPPEN complements upstream identifier-linkage systems, which use identity information available at the contributing site [7], [9], [11]. It provides a downstream image-based check after de-identified datasets have been combined. In this role, HAPPEN becomes part of the repository quality assurance workflow.

To support these applications, HAPPEN was designed for repositories that continue to grow and release updated data. First, it can reuse work from earlier iterations of dataset quality control: HAPPEN can output file fingerprints, preprocessing results, embeddings, and review decisions as an aggregated CSV file. Only new or changed scans need to be fingerprinted, preprocessed, and embedded, while these earlier review decisions can be reused for subsequent queries. Second, HAPPEN limits the number of pairs sent for human review. In the study repository, fewer than 10,000 candidate pairs required 4.5 hours of review. This kept the review workload practical despite being deployed on a large imaging collection. Future versions could reduce the number of preprocessing steps, incorporate them into the model, and distribute the processing across multiple machines.

This study also highlights several considerations that can be expanded upon in future work. First, monozygotic twins may be mistaken for the same individual based on anatomical similarity. Neither the genetic-reference criterion (PI_HAT > 0.9) nor image review reliably distinguishes these cases from identity duplication; future studies could incorporate pedigree or source identity records to resolve ambiguous cases. Second, Stage II has been evaluated only on T1-weighted brain MRI. Future studies can extend the model to T2-weighted MRI, FLAIR, diffusion MRI, and other imaging data. They can also explicitly evaluate the robustness of HAPPEN in scans with postoperative changes or lesions, over long follow-up intervals, and in the presence of substantial neurodegenerative change. Finally, the current study defined identity duplication as one person recorded under multiple subject identifiers. A complementary direction is to examine the reverse situation, in which one recorded identifier contains scans from more than one person. Future versions of HAPPEN could support this task by comparing scans within each identifier and flagging inconsistent groups for review.

## CREDIT AUTHORSHIP CONTRIBUTION STATEMENT

**Jiheng Li:** Conceptualization, Methodology, Software, Validation, Formal analysis, Investigation, Data curation, Visualization, Writing - original draft, Project administration. **Michael E. Kim:** Conceptualization, Data curation, Writing - review & editing, Supervision. **Trent M. Schwartz:** Supervision. **Yuhan Cui:** Investigation, Resources,

Data curation. **Gaurav Rudravaram:** Supervision. **Derek B. Archer:** Resources, Data curation, Funding acquisition. **Timothy J. Hohman:** Resources, Data curation, Funding acquisition. **Lori L. Beason-Held:** Resources. **Victoria L. Morgan:** Resources, Writing - review & editing, Funding acquisition. **Dario J. Englot:** Resources. **Angela L. Jefferson:** Resources, Funding acquisition. **Lianrui Zuo:** Supervision, Writing - review & editing. **Guray Erus:** Supervision, Project administration. **Christos Davatzikos:** Supervision, Project administration, Funding acquisition. **Bennett A. Landman:** Conceptualization, Writing - review & editing, Supervision, Project administration, Funding acquisition.

## DECLARATION OF COMPETING INTEREST

Timothy J. Hohman serves on the scientific advisory board of Circular Genomics and as an editor for Alzheimer's & Dementia and Alzheimer's & Dementia: Translational Research & Clinical Interventions. The remaining authors declare that they have no known competing financial interests or personal relationships that could have appeared to influence the work reported in this paper.

## CODE AND DATA AVAILABILITY

The HAPPEN source code and documentation are available at https://github.com/MASILab/HAPPEN. The versioned container image and launcher package are available on Zenodo at https://doi.org/10.5281/zenodo.21986366. The participant-level data analyzed in this study were obtained under the applicable repository access procedures and data-use agreements. The source imaging repositories are listed in Supplementary Table S1. The authors are not authorized to redistribute the aggregated study repository. Participant-level data from the external-site deployment remain under the control of the University of Pennsylvania and are not available from the authors.

## ACKNOWLEDGEMENTS

This work was supported by the Alzheimer's Disease Sequencing Project Phenotype Harmonization Consortium (ADSP-PHC), funded by the National Institute on Aging under awards U24 AG074855, U01 AG068057, and R01 AG059716. This work was also supported by NIH awards R01 EB017230, K01 AG073584, and K01 EB032898. This work was conducted in part using the resources of the Advanced Computing Center for Research and Education at Vanderbilt University in Nashville, Tennessee.

Data and/or research tools used in the preparation of this manuscript were obtained from the National Institute of Mental Health Data Archive (NDA). The NDA is a collaborative informatics system created by the National Institutes of Health to support and accelerate research in mental health. This manuscript reflects the views of the authors and may not reflect the opinions or views of the NIH or the researchers who submitted the original data to the NDA.

Data used in the preparation of this article were obtained from the Adolescent Brain Cognitive DevelopmentSM (ABCD) Study (https://abcdstudy.org), held in the NIMH Data Archive (NDA). This is a multisite, longitudinal study designed to recruit more than 10,000 children age 9-10 and follow them over 10 years into early adulthood. The ABCD Study® is supported by the National Institutes of Health and additional federal partners under award numbers U01DA041048, U01DA050989, U01DA051016, U01DA041022, U01DA051018, U01DA051037, U01DA050987, U01DA041174, U01DA041106, U01DA041117, U01DA041028, U01DA041134, U01DA050988, U01DA051039, U01DA041156, U01DA041025, U01DA041120, U01DA051038, U01DA041148, U01DA041093, U01DA041089, U24DA041123, U24DA041147. A full list of supporters is available at https://abcdstudy.org/federal-partners.html. A

listing of participating sites and a complete listing of the study investigators can be found at https://abcdstudy.org/consortium_members/. ABCD consortium investigators designed and implemented the study and/or provided data but did not necessarily participate in the analysis or writing of this report. This manuscript reflects the views of the authors and may not reflect the opinions or views of the NIH or ABCD consortium investigators.

VMAP data were collected by Vanderbilt Memory and Alzheimer's Center Investigators at Vanderbilt Health. VMAP began in 2012 with the goal of investigating vascular health and brain aging.

BLSA is a prospective cohort study with continuous enrollment that began in 1958. Comprehensive data from BLSA are available upon request by a proposal submission through the cohort website (www.blsa.nih.gov). The BLSA is supported by the Intramural Research Program of the National Institute on Aging, NIH.

The data contributed from the Wisconsin Registry for Alzheimer's Prevention was supported by NIA AG021155, AG027161, AG037639, and AG054047.

The Enhanced NKI-RS is a large cross-sectional sample of brain development, maturation and aging, which is currently funded by the NIMH (BRAINS R01MH094639-01; PI Milham) and Child Mind Institute (PI Milham) to characterize 1000 community-ascertained participants using state-of-the-art multiplex imaging-based resting state fMRI (R-fMRI) and diffusion tensor imaging (DTI), genetics, and a broad neurobehavioral phenotypic characterization protocol.

Imaging data and basic demographic information for the Nathan Kline Institute – Rockland Sample (NKI) dataset are freely available to download from https://fcon_1000.projects.nitrc.org/indi/enhanced/sharing_neuro.html. Phenotypic data are available upon request by filling out a data use agreement: https://fcon_1000.projects.nitrc.org/indi/enhanced/sharing_phenotypic.html.

The Bipolar & Schizophrenia Consortium for Parsing Intermediate Phenotypes (BSNIP1; 10.15154/tnzs-a323) dataset (R01MH078113-01, R01MH077852-01, R01MH077851-01, R01MH077945-01, R01MH077862-01) and its renewal (BSNIP2; 10.15154/8v3w-et72) (R01MH103368-01, R01MH103366-01) were collected with the aim to improve diagnosis and clinical management of psychosis by defining biologically based biotypes that better predict symptoms, course, and treatment response across major psychoses.

Previously acquired neuroimaging and associated participant data used in this study were obtained from the Knight Alzheimer Disease Research Center (Knight ADRC) at Washington University in St. Louis. Use of these data was reviewed and approved through the Knight ADRC resource request process and was conducted in accordance with Knight ADRC policies and applicable Washington University Human Research Protection Office (HRPO) requirements. Participant confidentiality was maintained throughout the study, and data were identified using ADRC participant identification numbers rather than participant names or initials. We gratefully acknowledge the Knight ADRC for providing the neuroimaging and associated data used in this study.

The resources provided by the Charles F. and Joanne Knight Alzheimer Disease Research Center (ADRC) for doi: 10.17632/sf7gxzdn63.2 were supported by the National Institutes of Health under award numbers P30AG066444, P01AG03991, P01AG026276, P30NS048056, R01AG078964, RF1AG044546, RF1AG071706, and RF1AG058501. We extend our sincere gratitude to the Knight ADRC research participants and their study partners for their invaluable time, contributions, and dedication to advancing this research. We also acknowledge the Knight ADRC Cores and Projects for their essential efforts in conducting, curating, and disseminating these datasets.

Data collection and sharing for the Alzheimer's Disease Neuroimaging Initiative (ADNI) is funded by the National Institute on Aging (National Institutes of Health Grant U19 AG024904). The grantee organization is the Northern California Institute for Research and Education. In the past, ADNI has also received funding from the National Institute of Biomedical Imaging and Bioengineering, the Canadian Institutes of Health Research, and private sector contributions through the Foundation for the National Institutes of Health (FNIH) including generous contributions from the following: AbbVie, Alzheimer's Association; Alzheimer's Drug Discovery Foundation; Araclon Biotech; BioClinica, Inc.; Biogen; Bristol-Myers Squibb Company; CereSpir, Inc.; Cogstate; Eisai Inc.; Elan Pharmaceuticals, Inc.; Eli Lilly and Company; EuroImmun; F. Hoffmann-La Roche Ltd and its affiliated company Genentech, Inc.; Fujirebio; GE Healthcare; IXICO Ltd.; Janssen Alzheimer Immunotherapy Research & Development, LLC.; Johnson & Johnson Pharmaceutical Research & Development LLC.; Lumosity; Lundbeck; Merck & Co., Inc.; Meso Scale Diagnostics, LLC.; NeuroRx Research; Neurotrack Technologies; Novartis Pharmaceuticals Corporation; Pfizer Inc.; Piramal Imaging; Servier; Takeda Pharmaceutical Company; and Transition Therapeutics.

The Amsterdam Open MRI Collection (AOMIC) is a collection of three datasets with multimodal (3T) MRI data including structural (T1-weighted), diffusion-weighted, and (resting-state and task-based) functional BOLD MRI data, as well as detailed demographics and psychometric variables from a large set of healthy participants. All raw data are publicly available from the OpenNeuro data sharing platform: ID1000: https://openneuro.org/datasets/ds003097, PIOP1: https://openneuro.org/datasets/ds002785, PIOP2: https://openneuro.org/datasets/ds002790/versions/2.0.0. We use version 1.2.1 for ID1000 and 2.0.0 for PIOP1 and PIOP2.

Data used in preparation of this article were derived from the BIOCARD study, supported by grant U19 – AG033655 from the National Institute on Aging. The BIOCARD study team did not participate in the analysis or writing of this report; however, they contributed to the design and implementation of the study. A listing of BIOCARD investigators can be found on the BIOCARD website (on the 'BIOCARD Data Access Procedures' page, 'Acknowledgement Agreement' document). A complete list of BIOCARD investigators can be found here: https://biocard.pathology.jhu.edu/our-team/.

The MORGAN dataset was supported in part by the National Institutes of Health [grant numbers R01 NS110130, R01 NS108445, R00 NS097618, R01 NS112252, R01 NS134625, and R01 NS146021].

The NACC database is funded by NIA/NIH Grant U24 AG072122. NACC data are contributed by the NIA-funded ADRCs: P30 AG062429 (PI James Brewer, MD, PhD), P30 AG066468 (PI Oscar Lopez, MD), P30 AG062421 (PI Bradley Hyman, MD, PhD), P30 AG066509 (PI Thomas Grabowski, MD), P30 AG066514 (PI Mary Sano, PhD), P30 AG066530 (PI Helena Chui, MD), P30 AG066507 (PI Marilyn Albert, PhD), P30 AG066444 (PI John Morris, MD), P30 AG066518 (PI Jeffrey Kaye, MD), P30 AG066512 (PI Thomas Wisniewski, MD), P30 AG066462 (PI Scott Small, MD), P30 AG072979 (PI David Wolk, MD), P30 AG072972 (PI Charles DeCarli, MD), P30 AG072976 (PI Andrew Saykin, PsyD), P30 AG072975 (PI David Bennett, MD), P30 AG072978 (PI Ann McKee, MD), P30 AG072977 (PI Robert Vassar, PhD), P30 AG066519 (PI Frank LaFerla, PhD), P30 AG062677 (PI Ronald Petersen, MD, PhD), P30 AG079280 (PI Eric Reiman, MD), P30 AG062422 (PI Gil Rabinovici, MD), P30 AG066511 (PI Allan Levey, MD, PhD), P30 AG072946 (PI Linda Van Eldik, PhD), P30 AG062715 (PI Sanjay Asthana, MD, FRCP), P30 AG072973 (PI Russell Swerdlow, MD), P30 AG066506 (PI Todd Golde, MD, PhD), P30 AG066508 (PI Stephen Strittmatter, MD, PhD), P30 AG066515 (PI Victor Henderson, MD, MS), P30 AG072947 (PI Suzanne Craft, PhD), P30 AG072931 (PI Henry Paulson, MD, PhD), P30 AG066546 (PI Sudha Seshadri, MD), P20 AG068024 (PI Erik Roberson, MD, PhD), P20 AG068053 (PI Justin Miller, PhD), P20 AG068077 (PI Gary Rosenberg, MD), P20

AG068082 (PI Angela Jefferson, PhD), P30 AG072958 (PI Heather Whitson, MD), P30 AG072959 (PI James Leverenz, MD).

Research reported in this publication as a part of the Health and Aging Brain Study: Health Disparities (HABS-HD) was supported by the National Institute on Aging of the National Institutes of Health under Award Numbers R01AG054073 and R01AG058533, R01AG070862, P41EB015922 and U19AG078109. The content is solely the responsibility of the authors and does not necessarily represent the official views of the National Institutes of Health.

Data contributed from MAP/ROS/MARS was supported by NIA R01AG017917, P30AG10161, P30AG072975, R01AG022018, R01AG056405, UH2NS100599, UH3NS100599, R01AG064233, R01AG15819 and R01AG067482, and the Illinois Department of Public Health (Alzheimer's Disease Research Fund). Data can be accessed at www.radc.rush.edu. More information about participant demographics and study information can be found here: https://www.rushu.rush.edu/research-rush-university/departmental-research/rush-alzheimers-disease-center/rush-alzheimers-disease-center-research/epidemiologic-research.

Data used in the preparation of this work were obtained from the International Consortium for Brain Mapping (ICBM) database (www.loni.usc.edu/ICBM). The ICBM project (Principal Investigator John Mazziotta, M.D., University of California, Los Angeles) is supported by the National Institute of Biomedical Imaging and BioEngineering. ICBM is the result of efforts of co-investigators from UCLA, Montreal Neurologic Institute, University of Texas at San Antonio, and the Institute of Medicine, Juelich/Heinrich Heine University - Germany. Data collection and sharing for this project was provided by the International Consortium for Brain Mapping (ICBM; Principal Investigator: John Mazziotta, MD, PhD). ICBM funding was provided by the National Institute of Biomedical Imaging and BioEngineering. ICBM data are disseminated by the Laboratory of Neuro Imaging at the University of Southern California.

The Dallas Lifespan Brain Study (DLBS) is a longitudinal multi-modal neuroimaging study of the aging mind, which was initiated in 2008 (referred to as Wave 1). Participants returned for two additional waves of data collection with an approximate interval of 4-5 years between waves. The DLBS protocol encompasses various imaging modalities, including structural MRI, diffusion MRI, and functional MRI, as well as comprehensive cognitive and psychosocial assessments. DLBS data can be downloaded from OpenNeuro here: https://openneuro.org/datasets/ds004856. Specifically, we use version 1.2.0.

The NACC database is funded by NIA/NIH Grant U24 AG072122. SCAN is a multi-institutional project that was funded as a U24 grant (AG067418) by the National Institute on Aging in May 2020. Data collected by SCAN and shared by NACC are contributed by the NIA-funded ADRCs as follows:

Arizona Alzheimer's Center - P30 AG072980 (PI: Eric Reiman, MD); R01 AG069453 (PI: Eric Reiman (contact), MD); P30 AG019610 (PI: Eric Reiman, MD); and the State of Arizona which provided additional funding supporting our center; Boston University - P30 AG013846 (PI Neil Kowall MD); Cleveland ADRC - P30 AG062428 (James Leverenz, MD); Cleveland Clinic, Las Vegas - P20AG068053; Columbia - P50 AG008702 (PI Scott Small MD); Duke/UNC ADRC - P30 AG072958; Emory University - P30AG066511 (PI Levey Allan, MD, PhD); Indiana University - R01 AG19771 (PI Andrew Saykin, PsyD); P30 AG10133 (PI Andrew Saykin, PsyD); P30 AG072976 (PI Andrew Saykin, PsyD); R01 AG061788 (PI Shannon Risacher, PhD); R01 AG053993 (PI Yu-Chien Wu, MD, PhD); U01 AG057195 (PI Liana Apostolova, MD); U19 AG063911 (PI Bradley Boeve, MD); and the Indiana University Department of Radiology and Imaging Sciences; Johns Hopkins - P30 AG066507 (PI Marilyn Albert,

PhD.); Mayo Clinic - P50 AG016574 (PI Ronald Petersen MD PhD); Mount Sinai - P30 AG066514 (PI Mary Sano, PhD); R01 AG054110 (PI Trey Hedden, PhD); R01 AG053509 (PI Trey Hedden, PhD); New York University - P30AG066512-01S2 (PI Thomas Wisniewski, MD); R01AG056031 (PI Ricardo Osorio, MD); R01AG056531 (PIs Ricardo Osorio, MD; Girardin Jean-Louis, PhD); Northwestern University - P30 AG013854 (PI Robert Vassar PhD); R01 AG045571 (PI Emily Rogalski, PhD); R56 AG045571, (PI Emily Rogalski, PhD); R01 AG067781, (PI Emily Rogalski, PhD); U19 AG073153, (PI Emily Rogalski, PhD); R01 DC008552, (M.-Marsel Mesulam, MD); R01 AG077444, (PIs M.-Marsel Mesulam, MD, Emily Rogalski, PhD); R01 NS075075 (PI Emily Rogalski, PhD); R01 AG056258 (PI Emily Rogalski, PhD); Oregon Health and Science University - P30 AG008017 (PI Jeffrey Kaye MD); R56 AG074321 (PI Jeffrey Kaye, MD); Rush University - P30 AG010161 (PI David Bennett MD); Stanford - P30AG066515; P50 AG047366 (PI Victor Henderson MD MS); University of Alabama, Birmingham - P20; University of California, Davis - P30 AG10129 (PI Charles DeCarli, MD); P30 AG072972 (PI Charles DeCarli, MD); University of California, Irvine - P50 AG016573 (PI Frank LaFerla PhD); University of California, San Diego - P30AG062429 (PI James Brewer, MD, PhD); University of California, San Francisco - P30 AG062422 (Rabinovici, Gil D., MD); University of Kansas - P30 AG035982 (Russell Swerdlow, MD); University of Kentucky - P30 AG028283-15S1 (PIs Linda Van Eldik, PhD and Brian Gold, PhD); University of Michigan ADRC - P30AG053760 (PI Henry Paulson, MD, PhD) P30AG072931 (PI Henry Paulson, MD, PhD) Cure Alzheimer's Fund 200775 - (PI Henry Paulson, MD, PhD) U19 NS120384 (PI Charles DeCarli, MD, University of Michigan Site PI Henry Paulson, MD, PhD) R01 AG068338 (MPI Bruno Giordani, PhD, Carol Persad, PhD, Yi Murphey, PhD) S10OD026738-01 (PI Douglas Noll, PhD) R01 AG058724 (PI Benjamin Hampstead, PhD) R35 AG072262 (PI Benjamin Hampstead, PhD) W81XWH2110743 (PI Benjamin Hampstead, PhD) R01 AG073235 (PI Nancy Chiaravalloti, University of Michigan Site PI Benjamin Hampstead, PhD) 1I01RX001534 (PI Benjamin Hampstead, PhD) IRX001381 (PI Benjamin Hampstead, PhD); University of New Mexico - P20 AG068077 (Gary Rosenberg, MD); University of Pennsylvania - State of PA project 2019NF4100087335 (PI David Wolk, MD); Rooney Family Research Fund (PI David Wolk, MD); R01 AG055005 (PI David Wolk, MD); University of Pittsburgh - P50 AG005133 (PI Oscar Lopez MD); University of Southern California - P50 AG005142 (PI Helena Chui MD); University of Washington - P50 AG005136 (PI Thomas Grabowski MD); University of Wisconsin - P50 AG033514 (PI Sanjay Asthana MD FRCP); Vanderbilt University - P20 AG068082; Wake Forest - P30AG072947 (PI Suzanne Craft, PhD); Washington University, St. Louis - P01 AG03991 (PI John Morris MD); P01 AG026276 (PI John Morris MD); P20 MH071616 (PI Dan Marcus); P30 AG066444 (PI John Morris MD); P30 NS098577 (PI Dan Marcus); R01 AG021910 (PI Randy Buckner); R01 AG043434 (PI Catherine Roe); R01 EB009352 (PI Dan Marcus); UL1 TR000448 (PI Brad Evanoff); U24 RR021382 (PI Bruce Rosen); Avid Radiopharmaceuticals / Eli Lilly; Yale - P50 AG047270 (PI Stephen Strittmatter MD PhD); R01AG052560 (MPI: Christopher van Dyck, MD; Richard Carson, PhD); R01AG062276 (PI: Christopher van Dyck, MD); 1Florida - P30AG066506-03 (PI Glenn Smith, PhD); P50 AG047266 (PI Todd Golde MD PhD)

HABS-HD comes from UNT Health Fort Worth, with the following primary investigators: Sid E O'Bryant (University of North Texas Health Science Center, Fort Worth, TX, USA); Kristine Yaffe (University of California San Francisco, San Francisco, CA, USA); Arthur Toga (University of Southern California, Los Angeles, CA, USA); Robert Rissman (University of Southern California, Los Angeles, CA, USA); and Leigh Johnson (University of North Texas Health Science Center, Fort Worth, TX, USA);

and the HABS-HD Investigators:

University of North Texas Health Science Center, Fort Worth, TX, USA: James R Hall, Melissa Petersen, Robert Barber, Fan Zhang, Rajesh Nandy, David Mason, Stephanie Large, Raul Vintimilla, Zhengyang Zhou, Rocky Vig, Nicole Phillips; University of Southern California, Los Angeles, CA, USA: Meredith Braskie, Kevin King, Michael Donohue, Rema Raman, Matthew Borzage, Yonggang Shi; Washington University in St. Louis, St. Louis, MO, USA: Beau Ances, Ganesh Babulal, Jorge Llibre-Guerra; University of Wisconsin–Madison, Madison, WI, USA: Bradley Christian, Amy Kind, Ozioma Okonkwo; Columbia University, New York, NY, USA: Badri Vardarajan, Joe Lee; Fordham University, New York, NY, USA: Monica Rivera Mindt; Georgetown University Medical Center, Washington, DC, USA: Amrita Cheema; The University of Texas Health Science Center, San Antonio, TX, USA: Raymond Palmer; University of Texas Southwestern Medical Center, Dallas, TX, USA: Roderick McColl; Rush University Medical Center, Chicago, IL, USA: Lisa Barnes; University of California Irvine School of Medicine, Irvine, CA, USA: Mark Mapstone; University of Pittsburgh School of Medicine, Pittsburgh, PA, USA: Annie Cohen; Wake Forest University School of Medicine, Winston-Salem, NC, USA: Michelle Mielke; Alzheimer's Association, Chicago, IL, USA: Carl Hill

## DECLARATION OF GENERATIVE AI AND AI-ASSISTED TECHNOLOGIES IN THE MANUSCRIPT PREPARATION PROCESS

During preparation of this manuscript, the authors used OpenAI ChatGPT to assist with language editing and stylistic refinement. All suggestions were reviewed and revised by the authors, who take full responsibility for the final manuscript.

## REFERENCES

[1] T. G. M. van Erp *et al.*, "Cortical Brain Abnormalities in 4474 Individuals With Schizophrenia and 5098 Control Subjects via the Enhancing Neuro Imaging Genetics Through Meta Analysis (ENIGMA) Consortium," *Biol. Psychiatry*, vol. 84, no. 9, 2018, doi: 10.1016/j.biopsych.2018.04.023.

[2] T. J. Littlejohns *et al.*, "The UK Biobank imaging enhancement of 100,000 participants: rationale, data collection, management and future directions," *Nat. Commun.*, vol. 11, no. 1, 2020, doi: 10.1038/s41467-020-15948-9.

[3] D. J. Hagler *et al.*, "Image processing and analysis methods for the Adolescent Brain Cognitive Development Study," *Neuroimage*, vol. 202, 2019, doi: 10.1016/j.neuroimage.2019.116091.

[4] E. E. Bron *et al.*, "Cross-cohort generalizability of deep and conventional machine learning for MRI-based diagnosis and prediction of Alzheimer's disease," *Neuroimage Clin.*, vol. 31, 2021, doi: 10.1016/j.nicl.2021.102712.

[5] Brain Imaging Data Structure Specification, "Data summary files," Brain Imaging Data Structure 1.11.0. Accessed: Feb. 16, 2026. [Online]. Available: https://bids-specification.readthedocs.io/en/stable/modality-agnostic-files/data-summary-files.html

[6] Open Access Series of Imaging Studies, "Datasets | Open Access Series of Imaging Studies (OASIS)." Accessed: Feb. 12, 2026. [Online]. Available: https://sites.wustl.edu/oasisbrains/datasets/

[7] UK Biobank, "UK Biobank Access Procedures (Version 2.1)." Accessed: Feb. 12, 2026. [Online]. Available: https://www.ukbiobank.ac.uk/wp-content/uploads/2025/01/Access-procedures.pdf

[8] UK Biobank, "Summary De-identification Protocol V2.1." Accessed: Feb. 12, 2026. [Online]. Available: https://www.ukbiobank.ac.uk/wp-content/uploads/2025/01/De-identification-protocol.pdf

[9] NIMH Data Archive, “NDA GUID Tool.” Accessed: Feb. 12, 2026. [Online]. Available: https://nda.nih.gov/s/guid/nda-guid.html

[10] S. B. Johnson *et al.*, “Using global unique identifiers to link autism collections,” *Journal of the American Medical Informatics Association*, vol. 17, no. 6, 2010, doi: 10.1136/jamia.2009.002063.

[11] National Alzheimer’s Coordinating Center, “Guidance for Management and Sharing of NACC Data (DSMP v1.0).” Accessed: Feb. 12, 2026. [Online]. Available: https://files.alz.washington.edu/documentation/nacc-data-sharing-guidance.pdf

[12] National Alzheimer’s Coordinating Center, “PTID to NACCID Map: Instructions.” Accessed: Feb. 12, 2026. [Online]. Available: https://files.alz.washington.edu/documentation/PTID-to-NACCID-map-instructions.pdf

[13] T. M. Shiovitz, C. S. Wilcox, L. Gevorgyan, and A. Shawkat, “CNS sites cooperate to detect duplicate subjects with a clinical trial subject registry,” *Innov. Clin. Neurosci.*, vol. 10, no. 2, 2013.

[14] T. M. Shiovitz *et al.*, “Mitigating the Effects of Nonadherence in Clinical Trials,” *J. Clin. Pharmacol.*, vol. 56, no. 9, 2016, doi: 10.1002/jcph.689.

[15] J. B. Poline *et al.*, “Data sharing in neuroimaging research,” *Front. Neuroinform.*, vol. 6, 2012, doi: 10.3389/fninf.2012.00009.

[16] S. E. Lazic, “The problem of pseudoreplication in neuroscientific studies: Is it affecting your analysis?,” *BMC Neurosci.*, vol. 11, 2010, doi: 10.1186/1471-2202-11-5.

[17] M. Rosenblatt, L. Tejavibulya, R. Jiang, S. Noble, and D. Scheinost, “Data leakage inflates prediction performance in connectome-based machine learning models,” *Nat. Commun.*, vol. 15, no. 1, 2024, doi: 10.1038/s41467-024-46150-w.

[18] M. Shim, S. H. Lee, and H. J. Hwang, “Inflated prediction accuracy of neuropsychiatric biomarkers caused by data leakage in feature selection,” *Sci. Rep.*, vol. 11, no. 1, 2021, doi: 10.1038/s41598-021-87157-3.

[19] E. Yagis *et al.*, “Effect of data leakage in brain MRI classification using 2D convolutional neural networks,” *Sci. Rep.*, vol. 11, no. 1, 2021, doi: 10.1038/s41598-021-01681-w.

[20] V. M. Young, S. Gates, L. Y. Garcia, and A. Salardini, “Data Leakage in Deep Learning for Alzheimer’s Disease Diagnosis: A Scoping Review of Methodological Rigor and Performance Inflation,” *Diagnostics*, vol. 15, no. 18, 2025, doi: 10.3390/diagnostics15182348.

[21] S. Kapoor and A. Narayanan, “Leakage and the reproducibility crisis in machine-learning-based science,” *Patterns*, vol. 4, no. 9, 2023, doi: 10.1016/j.patter.2023.100804.

[22] C. Wachinger, P. Golland, W. Kremen, B. Fischl, and M. Reuter, “BrainPrint: A discriminative characterization of brain morphology,” *Neuroimage*, vol. 109, 2015, doi: 10.1016/j.neuroimage.2015.01.032.

[23] L. Puglisi, F. Barkhof, D. C. Alexander, G. J. M. Parker, A. Eshaghi, and D. Ravì, “DeepBrainPrint: A Novel Contrastive Framework for Brain MRI Re-Identification,” *Proc. Mach. Learn. Res.*, vol. 227, 2023.

[24] L. Chauvin *et al.*, “Neuroimage signature from salient keypoints is highly specific to individuals and shared by close relatives,” *Neuroimage*, vol. 204, 2020, doi: 10.1016/j.neuroimage.2019.116208.

[25] The University of Melbourne Library, “Fixity checking.” Accessed: Feb. 17, 2026. [Online]. Available: https://library.unimelb.edu.au/Digital-Scholarship/digital-preservation/digital-preservation-guides/fixity-checking

[26] NIST, “FIPS 180-4: ‘Secure Hash Standard (SHS),’” 2015.

[27] "nipy/nibabel: 5.4.1," Zenodo. Accessed: Mar. 25, 2026. [Online]. Available: https://doi.org/10.5281/zenodo.18944560

[28] N. J. Tustison *et al.*, "N4ITK: Improved N3 bias correction," *IEEE Trans. Med. Imaging*, vol. 29, no. 6, 2010, doi: 10.1109/TMI.2010.2046908.

[29] L. Fisch *et al.*, "deepbet: Fast brain extraction of T1-weighted MRI using Convolutional Neural Networks," *Comput. Biol. Med.*, vol. 179, 2024, doi: 10.1016/j.compbiomed.2024.108845.

[30] B. B. Avants, N. J. Tustison, G. Song, P. A. Cook, A. Klein, and J. C. Gee, "A reproducible evaluation of ANTs similarity metric performance in brain image registration," *Neuroimage*, vol. 54, no. 3, 2011, doi: 10.1016/j.neuroimage.2010.09.025.

[31] V. Fonov, A. C. Evans, K. Botteron, C. R. Almli, R. C. McKinstry, and D. L. Collins, "Unbiased average age-appropriate atlases for pediatric studies," *Neuroimage*, vol. 54, no. 1, 2011, doi: 10.1016/j.neuroimage.2010.07.033.

[32] V. Fonov, A. Evans, R. McKinstry, C. Almli, and D. Collins, "Unbiased nonlinear average age-appropriate brain templates from birth to adulthood," *Neuroimage*, vol. 47, 2009, doi: 10.1016/s1053-8119(09)70884-5.

[33] D. L. Collins, A. P. Zijdenbos, W. F. C. Baare, and A. C. Evans, "ANIMAL+INSECT: Improved cortical structure segmentation," *Information Processing in Medical Imaging*, vol. 1613, 1999, doi: 10.1007/3-540-48714-x_16.

[34] P. Khosla *et al.*, "Supervised Contrastive Learning," *NeurIPS*, 2020, [Online]. Available: http://arxiv.org/abs/2004.11362

[35] J. Hu, L. Shen, S. Albanie, G. Sun, and E. Wu, "Squeeze-and-Excitation Networks," *IEEE Trans. Pattern Anal. Mach. Intell.*, vol. 42, no. 8, 2020, doi: 10.1109/TPAMI.2019.2913372.

[36] J. Johnson, M. Douze, and H. Jegou, "Billion-Scale Similarity Search with GPUs," *IEEE Trans. Big Data*, vol. 7, no. 3, 2021, doi: 10.1109/TBDATA.2019.2921572.

[37] M. E. Kim *et al.*, "Scalable quality control on processing of large diffusion-weighted and structural magnetic resonance imaging datasets," *PLoS One*, vol. 20, no. 8 August, 2025, doi: 10.1371/journal.pone.0327388.

[38] Vanderbilt Memory and Alzheimer's Center, "ADSP-PHC Data," Vanderbilt Memory and Alzheimer's Center Data Sharing Portal. Accessed: Feb. 22, 2026. [Online]. Available: https://vmacdata.org/adsp-phc/data

[39] L. Ferrucci, "The Baltimore Longitudinal Study of Aging (BLSA): A 50-year-long journey and plans for the future," *Journals of Gerontology - Series A Biological Sciences and Medical Sciences*, vol. 63, no. 12, 2008, doi: 10.1093/gerona/63.12.1416.

[40] A. L. Jefferson *et al.*, "The Vanderbilt Memory & Aging Project: Study Design and Baseline Cohort Overview," *Journal of Alzheimer's Disease*, vol. 52, no. 2, 2016, doi: 10.3233/JAD-150914.

[41] F. Pérez-García, R. Sparks, and S. Ourselin, "TorchIO: A Python library for efficient loading, preprocessing, augmentation and patch-based sampling of medical images in deep learning," *Comput. Methods Programs Biomed.*, vol. 208, 2021, doi: 10.1016/j.cmpb.2021.106236.

[42] B. J. Casey *et al.*, "The Adolescent Brain Cognitive Development (ABCD) study: Imaging acquisition across 21 sites," *Dev. Cogn. Neurosci.*, vol. 32, 2018, doi: 10.1016/j.dcn.2018.03.001.

[43] C. J. Weber *et al.*, "The Worldwide Alzheimer's Disease Neuroimaging Initiative: ADNI-3 updates and global perspectives," *Alzheimer's and Dementia: Translational Research and Clinical Interventions*, vol. 7, no. 1, 2021, doi: 10.1002/trc2.12226.

[44] M. Weiner *et al.*, "Effects of traumatic brain injury and posttraumatic stress disorder on development of Alzheimer's disease in Vietnam Veterans using the Alzheimer's Disease Neuroimaging Initiative: Preliminary report," *Alzheimer's and Dementia: Translational Research and Clinical Interventions*, vol. 3, no. 2, 2017, doi: 10.1016/j.trci.2017.02.005.

[45] L. Snoek, M. M. van der Miesen, T. Beemsterboer, A. van der Leij, A. Eigenhuis, and H. Steven Scholte, "The Amsterdam Open MRI Collection, a set of multimodal MRI datasets for individual difference analyses," *Sci. Data*, vol. 8, no. 1, 2021, doi: 10.1038/s41597-021-00870-6.

[46] M. I. Miller *et al.*, "Amygdalar atrophy in symptomatic Alzheimer's disease based on diffeomorphometry: The BIOCARD cohort," *Neurobiol. Aging*, vol. 36, no. S1, 2015, doi: 10.1016/j.neurobiolaging.2014.06.032.

[47] S. Kristian Hill *et al.*, "Neuropsychological impairments in schizophrenia and psychotic Bipolar disorder: Findings from the Bipolar-Schizophrenia Network on Intermediate Phenotypes (B-SNIP) study," *American Journal of Psychiatry*, vol. 170, no. 11, 2013, doi: 10.1176/appi.ajp.2013.12101298.

[48] D. A. Parker *et al.*, "Differentiating biomarker features and familial characteristics of B-SNIP psychosis Biotypes," *Transl. Psychiatry*, vol. 15, no. 1, 2025, doi: 10.1038/s41398-025-03501-5.

[49] D. C. Park *et al.*, "The Dallas Lifespan Brain Study: A Comprehensive Adult Lifespan Data Set of Brain and Cognitive Aging," *Scientific Data* , vol. 12, no. 1, 2025, doi: 10.1038/s41597-025-04847-7.

[50] S. E. O'Bryant *et al.*, "The health & aging brain among latino elders (Hable) study methods and participant characteristics," *Alzheimer's and Dementia: Diagnosis, Assessment and Disease Monitoring*, vol. 13, no. 1, 2021, doi: 10.1002/dad2.12202.

[51] J. C. Mazziotta *et al.*, "The myth of the normal, average human brain-The ICBM experience: (1) Subject screening and eligibility," *Neuroimage*, vol. 44, no. 3, 2009, doi: 10.1016/j.neuroimage.2008.07.062.

[52] "Indiana Alzheimer's Disease Research Center." Accessed: Aug. 21, 2026. [Online]. Available: https://medicine.iu.edu/research-centers/alzheimers

[53] National Institute on Aging, "KBASE2: Korean Brain Aging Study, Longitudinal Endophenotypes and Systems Biology," NIH RePORTER. Accessed: Mar. 16, 2026. [Online]. Available: https://reporter.nih.gov/search/Knos02QLrUWT9psRLHqdzw/project-details/10468028#clinical-studies

[54] M. S. Byun *et al.*, "Korean brain aging study for the early diagnosis and prediction of Alzheimer's disease: Methodology and baseline sample characteristics," *Psychiatry Investig.*, vol. 14, no. 6, 2017, doi: 10.4306/pi.2017.14.6.851.

[55] B. Akbarian *et al.*, "Association Between Postsurgical Functional Connectivity and Seizure Outcome in Patients With Temporal Lobe Epilepsy," *Neurology*, vol. 103, no. 7, 2024, doi: 10.1212/WNL.0000000000209816.

[56] A. Janson *et al.*, "Indirect structural changes and reduced controllability after temporal lobe epilepsy resection," *Epilepsia*, vol. 65, no. 3, 2024, doi: 10.1111/epi.17889.

[57] D. L. Beekly *et al.*, "The National Alzheimer's Coordinating Center (NACC) database: The uniform data set," *Alzheimer Dis. Assoc. Disord.*, vol. 21, no. 3, 2007, doi: 10.1097/WAD.0b013e318142774e.

[58] K. B. Nooner *et al.*, "The NKI-Rockland sample: A model for accelerating the pace of discovery science in psychiatry," *Front. Neurosci.*, vol. 6, 2012, doi: 10.3389/fnins.2012.00152.

[59] D. A. Bennett, J. A. Schneider, Z. Arvanitakis, and R. S. Wilson, "Overview and Findings from the Religious Orders Study," *Curr. Alzheimer Res.*, vol. 9, no. 6, 2012, doi: 10.2174/156720512801322573.

[60] D. A. Bennett, J. A. Schneider, A. S. Buchman, C. Mendes De Leon, J. L. Bienias, and R. S. Wilson, “The Rush Memory and Aging Project: study design and baseline characteristics of the study cohort,” *Neuroepidemiology*, vol. 25, no. 4, 2005, doi: 10.1159/000087446.
[61] L. L. Barnes, R. C. Shah, N. T. Aggarwal, D. A. Bennett, and J. A. Schneider, “The Minority Aging Research Study: Ongoing Efforts to Obtain Brain Donation in African Americans without Dementia,” *Curr. Alzheimer Res.*, vol. 9, no. 6, 2012, doi: 10.2174/156720512801322627.
[62] “SCAN.” Accessed: Mar. 16, 2026. [Online]. Available: https://scan.naccdata.org/
[63] D. A. Hoagey *et al.*, “The Knight Alzheimer Research Imaging dataset: a comprehensive multimodal resource for exploring aging, preclinical, and symptomatic Alzheimer disease pathology,” *Alzheimer’s & Dementia*, vol. 20, no. S2, 2024, doi: 10.1002/alz.086955.
[64] M. A. Sager, B. Hermann, and A. La Rue, “Middle-aged children of persons with Alzheimer’s disease: APOE genotypes and cognitive function in the Wisconsin Registry for Alzheimer’s Prevention,” in *Journal of Geriatric Psychiatry and Neurology*, 2005. doi: 10.1177/0891988705281882.

## SUPPLEMENTARY MATERIAL

### S1. Repository composition

Table S1 summarizes the 21 datasets in the study repository after exact-duplicate consolidation and initial scan-level quality assessment. It reports the number of recorded subjects in each dataset and the number with at least three eligible T1-weighted scans, the criterion used to form the primary longitudinal-rich model-development pool. BLSA and VMAP were reserved as withheld evaluation cohorts; the remaining 19 datasets formed the source population for model development and internal evaluation. Because WRAP and WRAPnew used the same subject identifiers, they were combined and are reported as WRAP.

**Table S1. Per-dataset composition after exact-duplicate consolidation and initial quality assessment.**

| Dataset | Recorded subjects | T1-weighted scans | Subjects with $\geq 3$ scans |
|---|---|---|---|
| ABCD [42] | 11,816 | 23,009 | 2,313 |
| ADNIDOD [43], [44] | 274 | 563 | 56 |
| ADNI [43] | 3,120 | 23,037 | 1,964 |
| AOMIC [45] | 1,370 | 3,210 | 913 |
| BIOCARD [46] | 212 | 508 | 98 |
| BLSA [39] | 1,191 | 4,271 | 627 |
| BSNIP1 [47] | 1,089 | 1,323 | 16 |
| BSNIP2 [48] | 742 | 867 | 18 |
| DLBS [49] | 464 | 967 | 180 |
| HABSHD [50] | 4,254 | 6,677 | 658 |
| ICBM [51] | 193 | 883 | 159 |
| Indiana [52] | 616 | 1,040 | 94 |
| KBASE [53], [54] | 1,228 | 2,032 | 178 |
| MORGAN [55], [56] | 256 | 414 | 38 |
| NACC [57] | 4,901 | 7,749 | 672 |
| NKI [58] | 1,324 | 2,311 | 321 |
| ROSMAPMARS [59], [60], [61] | 1,261 | 2,888 | 409 |
| SCAN [62] | 2,958 | 3,417 | 41 |
| VMAP [40] | 911 | 2,187 | 297 |
| WASHU [63] | 939 | 2,876 | 387 |
| WRAP [64] | 755 | 3,510 | 466 |

**S2. Stage II model development**

The Stage II encoder was developed with supervised contrastive learning. Repeated scans from the same repository subject record were treated as positive views, whereas scans from different records were treated as negatives. The preprocessing workflow described in the main manuscript was applied consistently during training and inference. PCA-based screening and data augmentation were used only during model development.

**S2.1 PCA near-duplicate screening**

After preprocessing and post-preprocessing quality assessment, the model-development dataset, BLSA, and VMAP were pooled for unsupervised near-duplicate screening before data partitioning. Incremental PCA was fitted to the preprocessed voxel data using a batch size of 352 and 176 output components. Cosine similarity in PCA space was used to identify highly similar pairs assigned to different repository subject records. One researcher reviewed the resulting candidates, and scans judged to be near duplicates were removed. This screening model was used only for data cleaning and was distinct from the PCA retrieval baseline, which was fitted later using only the training split.

**S2.2 Training-view construction and augmentation**

Augmentation was applied only to the training split. For subjects represented by three scans, one augmented view was generated from each scan, producing three original and three augmented views. For subjects represented by two scans, two augmented views were generated from each scan, producing two original and four augmented views. The resulting six views from each subject were treated as positives during supervised contrastive training, whereas views from different subjects were treated as negatives.

We used the TorchIO package for data augmentation [41]. The overall augmentation strategy can be divided into two categories: (1) intensity-based, including noise, gamma adjustment, blur, and bias field simulation; and (2) artifact-based, including ghosting and motion simulation. For artifact simulation, we used a mutually exclusive sampling strategy with a total probability of 0.15; once triggered, either ghosting or motion was selected with equal probability. Detailed augmentation parameters are provided in Table S2.

**Table S2. Data transformations used for supervised contrastive training.**

| Transformation | Type | Probability | Parameters |
|---|---|---|---|
| Gaussian noise | Intensity-based | 70% | standard deviation sampled in [0.0, 0.03] |
| Gamma adjustment | Intensity-based | 70% | log-gamma sampled in [-0.18, 0.18] |
| Gaussian blur | Intensity-based | 40% | blur std sampled in [0.0, 0.8] |
| Bias Field | Intensity-based | 50% | coefficient = 0.25, order = 3 |
| Ghosting artifact | Artifact-based | 7.5% | number of ghosts in {2, 3}; intensity in [0.2, 0.5] |
| Motion artifact | Artifact-based | 7.5% | rotation in $[0°, 4°]$, translation in [0, 2], 1 transform |
| Intensity clamping | Post-processing | 100% | clipped to [0, 1] |

**S2.3 Input representation, architecture, and objective**

After preprocessing, all scans were represented on the native 1-mm isotropic grid of the MNI152NLin2009cSym template. Each volume was center-cropped to 160 × 192 × 160 voxels and zero-padded when the available field of view was smaller. The 32 central axial slices were retained, yielding an input tensor of shape (B, 1, 160, 192, 32).

The encoder was a 3D ResNet-like network with squeeze-and-excitation channel attention [35]. A 7 × 7 × 7 convolutional stem was followed by 3D BatchNorm, ReLU, and 3D MaxPool. Four residual stages used block configuration (2, 2, 2, 2) and channel widths (64, 128, 256, 512). Each residual block contained two 3 × 3 × 3 convolutions and an SE module. The final feature map had shape (B, 512, 5, 6, 1). Adaptive global average pooling produced a 512-dimensional vector, which was passed through a two-layer projection MLP (512 to 1024 to 512) and LayerNorm. The resulting representation was L2-normalized and used both for supervised contrastive training and for near-duplicate retrieval.

**Supervised contrastive objective.** For an anchor embedding $z_i$, let $P(i)$ denote the set of indices of other views from the same subject in the batch, and let $A(i)$ denote all other indices in the batch. The loss is [34]:

$$\mathcal{L}_{\text{sup}} = \sum_i \frac{-1}{|P(i)|} \sum_{p \in P(i)} \log \frac{\exp(z_i \cdot z_p / \tau)}{\sum_{a \in A(i)} \exp(z_i \cdot z_a / \tau)}$$

$\tau$ is the temperature hyperparameter. We set $\tau = 0.1$ in all experiments.

**S2.4 Training hyperparameters**

Training is performed for 100 epochs without early stopping. We use the AdamW optimizer with initial learning rate of $1 \times 10^{-4}$, maximum learning rate of $1 \times 10^{-3}$, weight decay coefficient of $1 \times 10^{-2}$, and $\beta$ parameters of (0.9, 0.95). Weight decay was applied only to non-bias, non-normalization parameters to avoid regularization on BatchNorm or LayerNorm parameters. Learning-rate scheduling followed the OneCycle policy: the learning rate was linearly warmed up from $1 \times 10^{-4}$ to $1 \times 10^{-3}$ over the first 10 epochs and then decayed with cosine annealing to $1 \times 10^{-8}$ over the remaining 90 epochs. Training was performed with automatic mixed precision (AMP) together with GradScaler to improve training speed and numerical stability, with an initial scaling factor of $2^{14}$ and a growth interval of 2000.

**S3. Baseline and efficiency-benchmark implementation**

**3D SIFT-Rank baseline.** We evaluated 3D SIFT-Rank using the official Appearance+Geometry implementation of Chauvin et al. (commit d1b2558ee1b0bf8af3f336dffd6865f6379193aa). For 3D-SIFT-FS, the nonzero support of the corresponding FreeSurfer brainmask.mgz was applied to the native T1-weighted image. For 3D-SIFT-Shared, the method received the same preprocessed three-dimensional volume used to construct the HAPPEN input, without the subsequent 32-slice selection. Feature extraction and soft-Jaccard matching followed the official implementation using 30 nearest descriptors per query keypoint; all other settings were left at their defaults.

**Efficiency evaluation.** Efficiency was evaluated on a fixed 200-scan cohort using 32 workers, with an additional 10 scans used for warm-up. We report representation time, search time per query, and representation size. Preprocessing and training time were excluded. HAPPEN and PCA used flat FAISS search, whereas 3D SIFT-Rank used its official all-to-all matcher.